\documentclass[10pt,twocolumn,letterpaper]{article}

\usepackage{cvpr}
\usepackage{times}
\usepackage{epsfig}
\usepackage{graphicx}
\usepackage{amsmath}
\usepackage{amssymb}

\usepackage[breaklinks=true,bookmarks=false]{hyperref}

\cvprfinalcopy % *** Uncomment this line for the final submission

\def\cvprPaperID{5995} % *** Enter the CVPR Paper ID here

\usepackage{amsmath,amsfonts,bm}

\def\eqref#1{equation~\ref{#1}}
\def\1{\bm{1}}

\DeclareMathAlphabet{\mathsfit}{\encodingdefault}{\sfdefault}{m}{sl}
\SetMathAlphabet{\mathsfit}{bold}{\encodingdefault}{\sfdefault}{bx}{n}

\usepackage{hyperref}
\usepackage{url}
\usepackage{xspace} 
\usepackage{graphicx}
\usepackage{adjustbox}
\usepackage{array}
\newcolumntype{C}[1]{>{\centering\arraybackslash}p{#1}}

\usepackage{multirow}
\usepackage[skip=10pt]{caption}

\newcommand{\personacaptions}{Personality-Captions}
\newcommand{\ourmodel}{TransResNet}

\newcommand{\ShowTell}{\textsc{ShowTell}\xspace}
\newcommand{\ShowAttTell}{\textsc{ShowAttTell}\xspace}
\newcommand{\UpDown}{\textsc{UpDown}\xspace}

\title{Engaging Image Captioning via Personality}

\author{Kurt Shuster, Samuel Humeau, Hexiang Hu, Antoine Bordes, Jason Weston \\
Facebook AI Research \\
\texttt{\{kshuster,samueulhumeau,hexianghu,abordes,jase\}@fb.com} \\
}

\begin{document}

\maketitle
\thispagestyle{empty}
\begin{abstract}
Standard image captioning tasks 
such as COCO and Flickr30k
are factual, neutral in tone and (to a human) state the obvious (e.g., ``a man playing a guitar''). 
While such tasks are
useful to verify that a machine understands the content of
an image, they are not engaging to humans as captions.
With this in mind we define a new task,
\textsc{\personacaptions}, where the goal is to be as engaging to humans as possible by incorporating controllable style and personality traits.
We collect and release a large dataset of 241,858 of such captions conditioned over 215 possible traits.
We build models that combine existing work from (i) sentence representations \cite{training_millions} with Transformers trained on 1.7 billion dialogue examples; and (ii)  image representations \cite{uru} with ResNets trained on 3.5 billion social media images.  We obtain state-of-the-art performance on Flickr30k and COCO, and strong performance on our new task.
Finally, online evaluations validate that our task and models are engaging to humans, with our best model close to human performance.
\end{abstract}

\section{Introduction}
%Outlines  https://fb.quip.com/n1ISANTDhLri

% If we want machines to communicate with humans,
% it is vital they can capture our %\footnote{The use of ``our'' here assumes that you are a human reading this.} 
% interest, which means spanning both the ability to understand and the ability to be engaging, in particular to display emotion and personality as well as conversational function 
% \cite{costa2010emotional,jay2007filling,jonczyk2016affect,scheutz2006utility,kampman2019adapting}.
% %\cite{cassell2000nudge}. 

If we want machines to communicate with humans,
%If we want machines to sustain discussions with humans,
they must be able to capture our interest, which means spanning both the ability to understand and the ability to be engaging, in particular to display emotion and personality as well as conversational function 
\cite{jay2007filling,jonczyk2016affect,scheutz2006utility,kampman2019adapting}.
Consider for example an online conversational agent or robot that can both {perceive images} and {speak} -- the afforementioned capabilities would be expected from a good conversationalist.
%This is particularly sensible in the case of non-goal oriented conversational agents (chitchat agents, or chatbots) where those capabilities seem expected from a good conversationalist.
%\cite{cassell2000nudge}. 

%Visual support is increasingly becoming important for communication. Whatsapp, one of the leading messaging applications worldwide reports that in 2017, 4.5 billion images are shared per day, 8\% of the total number of messages sent\footnote{\tiny\url{https://blog.whatsapp.com/10000631/Connecting-One-Billion-Users-Every-Day/}}. In order to develop engaging conversational agents, it seems promising to allow them to  comment on images naturally as humans do.
%comment on a picture and discuss naturally about it.

Communication grounded in images is naturally engaging
to humans \cite{hu2014we}.
For example, 4.5 billion images are shared per day  on the messaging platform WhatsApp\footnote{\tiny\url{https://blog.whatsapp.com/10000631/Connecting-One-Billion-Users-Every-Day/}} where users then naturally discuss those images. In order to develop engaging conversational agents, it thus seems promising to allow them to 
comment on images naturally as humans do.
Yet the majority of studies in the 
research community have so far focused on function only: standard image captioning \cite{pan2004automatic} requires the machine to generate a sentence which factually describes the elements of the scene in a neutral tone.
Similarly, visual question answering \cite{antol2015vqa} and visual dialogue \cite{das2017visual} require the machine to answer factual questions about the contents of the image, either in single turn or dialogue form. 
They assess whether the machine can perform basic perception over the image which humans take for granted. Hence, they are useful for developing models that understand content, but are not useful as an end application unless the human cannot see the image, e.g. due to visual impairment \cite{gurari2018vizwiz}. 

Standard  image captioning tasks 
 simply state the obvious, and are not considered engaging captions by humans. 
For example, in the COCO \cite{chen2015microsoft} and Flickr30k \cite{young2014image}
  tasks, some examples of captions include ``a large bus sitting next to a very tall building'' and
``a butcher cutting an animal to sell'', which describe the contents of those images in a personality-free, factual manner. However, humans consider engaging and effective captions ones that ``avoid stating the obvious'', as %can be seen from
shown by advice to human captioners outside of vision research.\footnote{\tiny\url{https://www.photoup.net/how-to-write-more-engaging-photo-captions/}} For example, ``If the bride and groom are smiling at each other, don't write that they are smiling at each other. The photo already visually shows what the subject is doing. Rephrase the caption to reflect the story behind the image''. Moreover, it is considered that ``conversational language works best. Write the caption as though you are talking to a family member or friend''.\footnote{\tiny\url{https://www.poynter.org/news/6-tips-writing-photo-captions}} 
These instructions %for human captioners 
to engage human readers 
%are in stark contrast to standard captioning datasets.
seem to be in direct opposition to standard captioning datasets.
%and finally you may want to ``direct the reader to a detail in the photo that you want them to focus on''. %These seem in direct opposition to the way captions are collected in e.g. COCO where the instructions were 
%``do not describe unimportant details'', ``do not describe what a person might say''

In this work we focus on image captioning that is engaging for humans by incorporating personality. As no large  dataset exists that covers the range of human personalities, we build and release a new dataset, \textsc{\personacaptions}, with  241,858 captions, each conditioned on one of 215 different possible personality traits.
We show that such captions are 
far more engaging %to humans 
than traditional ones.
% than traditional ones, and (ii) compared to simply 
% asking humans
% for an engaging caption {\em without conditioning} on a given trait our captions are more controllable, diverse and give higher engagement scores for certain traits. 

\begin{figure*}
\centering
\def\arraystretch{1.00}
\begin{tabular}{|c|l| }
\hline 
\multirow{ 9}{*}{\includegraphics[width=15.9ex]{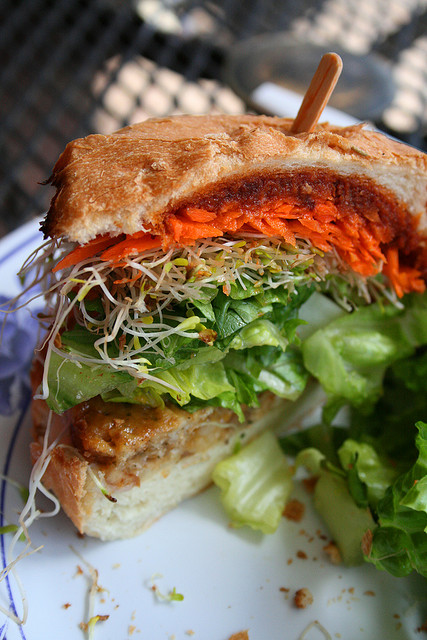}}
& \small{\textbf{Standard captioning output:} A plate with a sandwich and salad on it.} \rule{0pt}{2.7ex}\\
& \small{\textbf{Our model with different personality traits (215 possible traits, not all shown here):} }\\
& \small{\textit{Sweet}} \small{$\qquad\qquad$That is a lovely sandwich.} \\
& \small{\textit{Dramatic}} \small{$\qquad\,\,\,\,$This sandwich looks so delicious! My goodness!} \\
& \small{\textit{Anxious}} \small{$\qquad\quad\,\,$I'm afraid this might make me sick if I eat it.} \\
& \small{\textit{Sympathetic}} \small{$\quad\enspace\,$I feel so bad for that carrot, about to be consumed.}\\
& \small{\textit{Arrogant}} \small{$\qquad\enspace\,\,\,$I make better food than this} \\
& \small{\textit{Optimistic}} \small{$\qquad\,\,\,$It will taste positively wonderful!} \\
& \small{\textit{Money-minded}} \small{$\enspace\,$I would totally pay \$100 for this plate.} \\[0.3em]
\hline
\end{tabular}
\caption{Our \ourmodel~model compared to a standard image captioning model on the same image conditioned on various personality traits. Our model is trained on the new \textsc{\personacaptions} dataset which covers 215 different personality traits. The standard captioning system used for comparison is the best  COCO \UpDown model described in Section \ref{genmodeldesc}.
\label{fig:main_example}
}
\end{figure*}

We then develop model architectures that can simultaneously understand image content and provide engaging captions for humans. 
To build strong models, we consider both retrieval and generative\footnote{\tiny{"Generative" here refers to a model that generates a caption word-by-word as opposed to a retrieval model.}} 
%it from a set.}}
variants,
and  leverage
 state-of-the-art modules from both the vision and language domains.
 For image representations, we employ the work of \cite{uru} that uses
a ResNeXt architecture trained on 3.5 billion social media images which we apply to both. For text, we use a Transformer sentence representation following \cite{training_millions} trained on 1.7 billion dialogue examples. Our generative model gives a new state-of-the-art on COCO caption generation, and our retrieval architecture, \ourmodel, yields the highest known R@1 score on the Flickr30k dataset.
To make the models more engaging to humans,
we then adapt those same architectures to the 
 \textsc{\personacaptions} task by conditioning the input image on the given personality traits, giving strong performance on our new task,
 see Figure~\ref{fig:main_example}. In particular, when compared
to human captions, annotators  preferred our retrieval model's captions over human ones
49.5\% of the time -- very close to human performance.
%where the difference is not statistically significant.
Our task is however a challenge for generative models which
succeed on COCO, but fail on our task. 
We believe future work should address this important open  problem.

\section{Related Work}

A large body of work has
focused on developing image captioning datasets and  models that work on
them. In this paper we also perform experiments on the COCO \cite{chen2015microsoft} and Flickr30k \cite{young2014image} datasets, comparing to a range of models, including both generative models such as in \cite{vinyals2015show,xu2015show,anderson2017bottom} and retrieval based such as in 
\cite{DBLP:journals/corr/abs-1711-06420,DBLP:journals/corr/FaghriFKF17,DBLP:journals/corr/NamHK16}. These setups measure the ability of models to understand the content of an image, but do not address more natural human communication.

A number of works have tried to induce more engaging captions for human readers.
One area of study is to make the caption personalized to the
reader, e.g. by using user level features such as location and age \cite{denton2015user} or knowledge of the reader's active vocabulary \cite{8100164}. Our work does not address this issue.
Another research direction is  to attempt to produce amusing captions
either through wordplay (puns) \cite{chandrasekaran2017punny}
or training on data from humour websites \cite{yoshida2018neural}. Our work focuses on a general set of personality traits, not on humour.
Finally, closer to our work are approaches that attempt to model the style of the caption.
Some methods have tried to learn style in an unsupervised fashion, as a supervised dataset like we have built in this work was not available. As a result, evaluation was more challenging in those works, see e.g.
\cite{mathews2018semstyle}. Others such as \cite{you2018image} have used small datasets like SentiCap \cite{mathews2016senticap} with
 $\sim$800 images to inject sentiment into captions. \cite{gan2017stylenet} collect a somewhat bigger dataset with 10,000 images, FlickrStyle10K, but only covers two types of style (romantic and humorous).
In contrast, our models are trained on the
 \textsc{\personacaptions} dataset that has 215 traits and $\sim$200,000 images.

\begin{table*}[h!]

\begin{center}
\begin{small}
\begin{tabular}{|l|ccc|c|cc|cc|}
\hline 
Type & \multicolumn{4}{c|}{Datasets With Personality} & \multicolumn{4}{c|}{Datasets Without Personality} \\
 \hline
Dataset & \multicolumn{3}{c|}{\personacaptions} & FlickrStyle10K  & \multicolumn{2}{c|}{COCO}  & \multicolumn{2}{c|}{Flickr30k}  \\
\hline
Split & train & valid & test & train & train & valid & train & valid \\
\hline
Number of Images &    186,858  &  5,000  & 10,000 & 7000 & 82783 & 40504  & 29000 & 1014 \\
\hline
Number of Captions &    186,858  &  5,000 & 50,000 & 14000 & {414113} & 202654  & 145000 & 5070\\
\hline
Number of Personality Types  &  {215} & {215} & {215} & 2 & None & None   & None & None \\
\hline
Vocabulary Size & {33641}  & 5460 & 16655 & 8889 & 23776 & 17724  & 17920 & 4283 \\
\hline
Average Tokens per Caption   & 11.2 & 10.9 & 11.1 & {14.51} & 11.3 & 11.3  & 13.53 & 13.74 \\
\hline
\end{tabular}
\caption{\textsc{\personacaptions} dataset statistics compared to other captioning datasets. %To our knowledge, there is no other dataset incorporating personality/style traits of comparable size.
%To the best of our knowledge the authors of FlickRStyle10k haven't released the validation and test set.
\label{table:data_stats}
\vspace{-5mm}
}
\end{small}
\end{center}
\end{table*}

 Our work can also be linked to the more general area of human communication, separate from just factual captioning, in particular image grounded conversations between humans
 \cite{DBLP:journals/corr/MostafazadehBDG17} 
 or dialogue in general where displaying personality is 
 important \cite{zhang2018personalizing}.
 In those tasks, simple word overlap based automatic 
 metrics are shown to perform weakly
 \cite{liu2016not} due to the 
 intrinsically more diverse outputs in the tasks.
 As in those domains, we thus also perform human
 evaluations in this work to measure the 
 engagingness of our setup and models.
 
 In terms of modeling, image captioning performance is clearly boosted with any advancements
 in image or text encoders, particularly the former.
In this work we make use of the latest advancements in
image encoding by using the work of \cite{uru}
which provides state-of-the-art performance on ImagenNet image classification, but has so far not been applied to captioning.
For text encoding we use the latest advances in attention-based representations using Transformers \cite{vaswani2017attention}; in particular, their use in retrieval models for dialogue by large-scale pretraining \cite{training_millions} is adapted here for our captioning tasks.

\section{\personacaptions}

The \textsc{\personacaptions} dataset is a large
collection of (image, personality trait, caption) triples that we collected using crowd-workers, publicly available at {\small \url{http://parl.ai/projects/personality_captions}}.
%and will be made publicly available upon acceptance. 

\paragraph{Personality traits} A large number of studies are dedicated to produce a model of the personality of an individual \cite{Jacques2018FirstIA}, such as the Big-Five \cite{bigfive}, the Big-Two \cite{bigfive} and 16PF among others \cite{16pf}. Those models usually project personality in a low dimension space, for instance the Big-Five describes a personality by weighting openness to experience, conscientiousness, extraversion, agreeableness and neuroticism. However such a description is not well adapted to a crowdsourced data collection task, where labelers are not familiar with those models. We found it clearer to use a single descriptor as a ``personality trait''
 (e.g. ``sweet", ``skeptical", ``solemn", etc.).  We considered 215 possible personality traits which were constructed by selecting a subset from a curated list of 638 traits\footnote{\tiny\url{http://ideonomy.mit.edu/essays/traits.html}}
that we deemed suitable for our captioning task. 
%We did not select traits that seemed redundant or would not have impact on the comments created.
The traits are categorized into three classes:
positive (e.g., sweet, happy, eloquent, humble, perceptive, witty),
neutral (e.g., old-fashioned, skeptical, solemn, questioning) and negative (e.g., anxious, childish, critical, fickle).
Examples of traits that we did not use are allocentric, insouciant, flexible, earthy and invisible, due to the difficulty of their interpretation with respect to captioning an image.

\paragraph{Data collection} We use a randomly selected set of the images from the YFCC100M Dataset\footnote{\tiny\url{https://multimediacommons.wordpress.com/yfcc100m-core-dataset/}; \tiny\cite{Thomee:2016:YND:2886013.2812802}} to build our training, validation and test sets, selecting for each chosen image a random personality trait, drawn uniformly from our list. The captions are written by a large number of crowd-workers, with the annotation task distributed among them.
Test examples have 5 captions per image in order to compute multi-reference automatic evaluations such as BLEU.

%With the given image and persona, the annotator was asked to write an engaging caption while emulating the persona. 

In each annotation round, an annotator is  shown an image along with a trait.
The annotators are then asked to
%We then desire the annotators to 
write an engaging utterance for the image in the context of the personality trait.
Specifically, they are told to ``write a comment {\em in the context of your given personality trait}\dots about an image that someone else would find engaging''. Note we do not use the word ``caption'' in these instructions %but rather than the word ``comment'' 
because we felt it would be clearer to crowdworkers  of our intent: not many humans have experience writing captions and they may misinterpret the word to mean a factual netural statement, whereas they have experience writing personality-based engaging comments.
We thus aim to illicit more natural utterances that humans are used to writing. In this paper we refer to these labels as \textsc{Personality-Captions}.
% The goal of this task is to write something about an image that someone else would find engaging.

The captions are constrained to include at least three words. 
It was emphasized that the personality trait describes a trait of the author of the caption, not properties of the content of the image. 
They were also instructed not to use the personality trait word itself in their caption. For quality control, crowdworkers were manually monitored and removed for poor performance.
See Figure \ref{annotation_setup} in the appendix for more details of the exact instructions given to annotators.
The final dataset statistics are given in Table \ref{table:data_stats} and compared to the largest dataset we are aware of that also has personality based captions, FlickrStyle10k, which is significantly smaller in terms of images, examples and number of personalities. We also show standard captioning datasets COCO and Flickr30k for reference.

\section{Models}

We consider two classes of models for caption prediction: retrieval models and generative models. Retrieval models produce a caption by considering any caption in the training set as a possible candidate response. Generative models generate word-by-word novel sentences conditioned on the image and personality trait (using a beam).
Both approaches require an image encoder. %and text en(de)coders.
%Both approaches require both image and text 
%modules.

\subsection{Image Encoders}
\label{imagefeats}

We build both types of model on top of pretrained image features, and compare the performance of two types of image encoders. The first is a residual network with 152 layers described in \cite{resnet} trained on Imagenet \cite{imagenet} to classify images among 1000 classes, which we refer to in the rest of the paper as {\em ResNet152} features. We used the implementation provided in the torchvision project \cite{torchvision}. The second is a ResNeXt $32\times48$d \cite{resnext} trained on 3.5 billion Instagram pictures following the procedure described by \cite{uru}, which we refer to in the rest of the paper as {\em ResNeXt-IG-3.5B}. The authors provided the weights of their trained model to us. Both networks embed images in a 2048-dimensional vector which is the input for most of our models. In some of the caption generation models that make use of attention, we keep the spatial extent of the features by adapting activation before the last average pooling layer, and thus extract features with $7 \times 7 \times 2048$ dimensions.

\subsection{Caption generation models}
\label{genmodeldesc}
% \paragraph{Model Details.} 
We re-implemented three widely used previous/current state-of-the-art image captioning methods: % as representative approaches: 
\ShowTell~\cite{vinyals2015show}, \ShowAttTell~\cite{xu2015show} and \UpDown~\cite{anderson2017bottom}. %We explain in details about each model in the following part of this section.

\paragraph{Image and Personality Encoders} The image representation \(r_I\)  is extracted using the aforementioned 
image encoder. For the \ShowTell model, the 2048-dimensional outputs of image encoder is used. For the \ShowAttTell and \UpDown models, we keep the spatial extent and use the $7 \times 7 \times 2048$ dimensional outputs of image encoder.
In all cases, the image features are ultimately reduced to a vector of dimension 512. 
In the \ShowTell model, a linear projection is applied to do so. In both the \ShowAttTell and \UpDown models, the image features are first linearly reduced to a tensor of $7 \times 7 \times 512$ dimensions with a $1 \times 1$ convolution layer. Then the attention mechanism is used to weighted combine image features along its $7 \times 7$ spatial extent,
into a vector of dimension 512. 
In the cases where personality traits are used, each personality trait is embedded by a vector of dimension 512, akin to a word embedding, giving a $215 \times 512$ matrix of weights to learn for {\sc \personacaptions}. The personality embedding is then input to the LSTM caption decoders, through concatenating with the input word vectors at each decoding step.

\paragraph{Caption Decoders}  In \ShowTell, similar to \cite{vinyals2015show}, the dimensionality reduced image features are used as the first input word to a LSTM model to generate the output caption sequence. In \ShowAttTell, while the overall architecture is similar to~\cite{xu2015show}, we adopt the modification suggested by~\cite{rennie2017self} and input the attention-derived image features to the cell node of the LSTM. Finally, we use the \UpDown model exactly as described in~\cite{anderson2017bottom}. The key difference to \ShowAttTell is that two LSTM instead of one are used, of which one is responsible for generating the attention weight and the other is responsible of generating the caption. In all above models, the word vector of the previously predicted word (concatenated with personality embedding when applicable) is input to the LSTM caption decoder to predict the current word, at each caption decoding step.

\paragraph{Training and Inference} We perform a two-stage training strategy to train such caption generation models as proposed by~\cite{rennie2017self}. In the first stage,  we train the model to optimize the standard cross-entropy loss. In the second stage, we perform policy gradient with \textsc{reinforce} to optimize the non-differentiable reward function (CIDEr score in our case). During inference, we apply beam search (beam size=2) to decode the caption.

\subsection{Caption retrieval models}
\label{retrievalmodels}

%We also explore retrieval based  models.
We define a simple yet powerful retrieval 
architecture, named
\ourmodel. It works by projecting the image, personality, and caption in the same space \(S\) using image, personality, and text encoders.

%Similar to \sh{Insert relevant citations here}, 
%We also explore retrieval based image captioning models.
%Our retrieval based captioning models work by projecting the image, personality, and caption in the same space \(S\) using image, personality, and text encoders. %We define a simple yet powerful architecture, named
%\ourmodel,  which is detailed in Figure \ref{transresnet_schema}. 

\paragraph{Image and Personality Encoders} The representation \(r_I\) of an image \(I\) is obtained by using the 2048-dimensional output of the image encoder described in Sec. \ref{imagefeats} as input to a multi-layer perceptron with ReLU activation units and a final layer of 500 dimensions. To take advantage of personality traits in the
\textsc{\personacaptions} task, we embed each trait %to a 500-dimensional vector 
to obtain its representation \(r_P\) $\in \mathbb{R}^{500}$. Image and personality representations are then summed.

\paragraph{Caption Encoders} 
\vbox{
Each caption is encoded into a vector \(r_C\) of the same size using a Transformer architecture \cite{vaswani2017attention}, followed by a two layer perceptron.
We consider a Transformer architecture with 4 layers, 300 hidden units and 6 attention heads.
%We try two sizes of Transformer: a larger architecture (4 layers, 300 hidden units, 6 attention heads) and a smaller one (2 layers, 300 hidden units, 4 attention heads).
We either train from scratch, pretrain only the word embeddings, i.e. where we initialize word vectors trained using fastText \cite{bojanowski2016enriching} trained on Wikipedia, or pretrain the entire encoder. 
For the latter, we follow the setup described in \cite{training_millions}: we train two encoders on a next-utterance retrieval task on a dataset of dialogs containing 1.7 billion pairs of utterances, where one encodes the context and another the candidates for the next utterance, their dot product indicates the degree of match, and they are  
 trained with negative log-likelihood and $k$-negative sampling.
%Both are trained such that the dot product between the representations of the context and the right candidate is maximized.  
We then initialize our system using the weights of the candidate encoder only, and then train on our task.
%Caption training is done in two steps: we first train the upper layers while the transformer remains frozen. When maximum performance is obtained on the validation set we unfreeze the weights of the transformer.

\if 0
We try three variations:
\begin{itemize}
  \item The transformer is pretrained. We follow the setup described in \cite{training_millions}: we train 2 encoders on a next-utterance retrieval task on a dataset of dialogs containing 1.7 billion pairs of utterances. In this setup one of the transformer encodes the context and another encodes the candidates for the next utterance. We initialize our system using the weights of the candidates encoder. We used 4 layers, 6 attention heads and 300 hidden units. Training is done in two times, we first train the upper layers while the transformer remains frozen. Only when maximum performance is obtained on the validation set we release the weights of the transformer.
  \item Only word embeddings are pre-trained. We initialized them with word vectors trained using fastText \cite{bojanowski2016enriching} on wikipedia. In this case we found out that a much smaller architecture produces best results and used only 2 layers in the transformer with 300 hidden units and 4 attention heads.
  \item No pre-training at all. The same architecture is used for the transformer: 2 layers, 300 hidden units, 4 attention heads.
\end{itemize}
\fi
}

For comparison, we 
also consider a simple bag-of-words encoder (pretrained or not). In this case, \(r_C\) $\in \mathbb{R}^{300}$ is the sum of the word embeddings of the caption. 

In each case, given an input image and personality trait \((I, P)\) and a candidate caption \(C\), the score of the final combination is then computed as the following dot product:
\(s(I, P, C) = (r_I + r_P) \cdot r_C\).

\begin{figure*}[t!]
\begin{center}
\includegraphics[trim={0 1mm 0 2mm}, clip, width=0.92\textwidth]{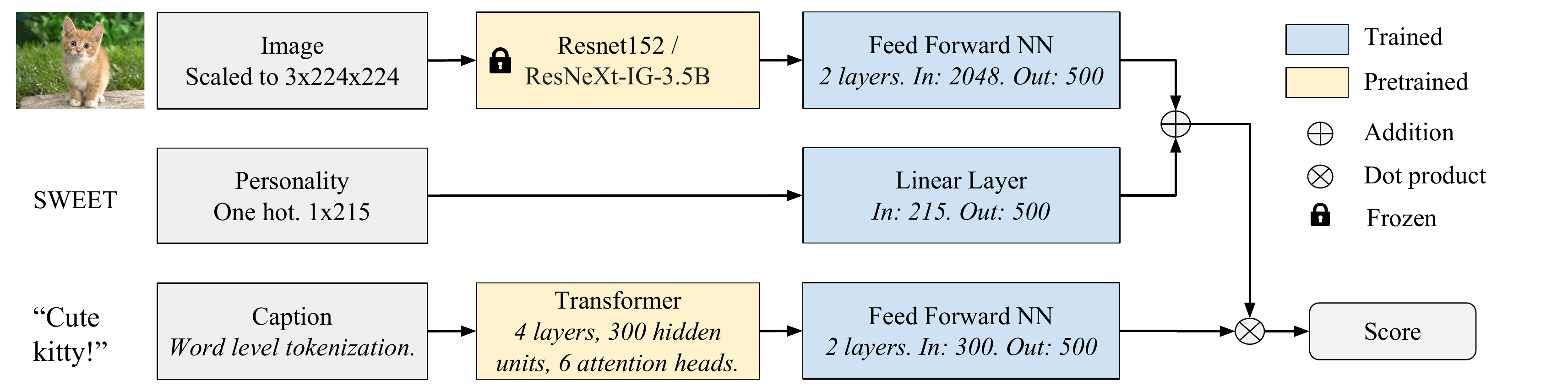}
\caption{Our architecture \ourmodel, used for our retrieval models.}
\label{transresnet_schema}
\label{retrieval_model}
\end{center}
\vspace*{-2mm}
\end{figure*}

\paragraph{Training and Inference} 
Given a pair \(I, P\), and a set of candidates \((c_1, .., c_N)\), at inference time the predicted caption is the candidate \(c_i\) that maximizes the score \(s(I, P, c_i)\). At training time we pass  a set of scores through a softmax and train to maximize the log-likelihood of the correct responses. We use mini-batches of 500 training examples; for each example, we use the captions of the other elements of the batch as negatives. 
%We refer to our retrieval model in the rest of the paper as \ourmodel. The overall architecture is detailed in Figure \ref{transresnet_schema}.
Our overall \ourmodel~architecture is detailed in Figure \ref{transresnet_schema}.

\section{Experiments}

We first test our architectures on traditional caption datasets to assess their ability to factually describe the contents of images
in a neutral tone. 
We then apply the same architectures to \textsc{\personacaptions}~to assess their ability to produce engaging captions conditioned on personality. The latter is tested with both automatic metrics and human
evaluation of both engagingness and fit.
%Finally we submit the different models to human evaluation and compare the engagingness of the captions they produce.

\subsection{Automatic evaluation on Traditional Captions}% Datasets}

\paragraph{Generative Models}

For our generative models, we test the quality of our implementations of existing models (\ShowTell, \ShowAttTell and \UpDown) as well as the quality of our image encoders, 
%where we compare 
ResNet152 and  ResNeXt-IG-3.5B. We report performance on the COCO caption dataset~\cite{lin2014microsoft}. We evaluate BLEU~\cite{papineni2002bleu}, ROUGE-L~\cite{lin2004rouge}, CIDEr~\cite{vedantam2015cider} and SPICE~\cite{anderson2016spice} and compare models' performances to state-of-the-art models under the setting of \cite{karpathy2015deep}.  We provide additional
ablations in Appendix \ref{asec:compare_coco}.

The results are shown in Table~\ref{autometric_coco_caption}. Models trained with ResNeXt-IG-3.5B features consistently outperform their counterparts with ResNet152 features, %which
demonstrating the effectiveness of ResNeXt-IG-3.5B  beyond the original image classification and detection results in \cite{uru}.
More importantly, our best model (\UpDown)  either outperforms or is competitive with state-of-the-art single model performance~\cite{anderson2017bottom} across most metrics (especially CIDEr). %Note that~\cite{anderson2017bottom} is using a ResNet based Faster RCNN~\cite{ren2015faster} pretrained on the Visual Genome~\cite{krishna2017visual} and adopted a similar two-stage training strategy as ours.  
%We train our model on both the training and the unused validation data of COCO captions as it is customary, see~\cite{rennie2017self,anderson2017bottom}.

%\sh{
%(could be also be merged with the next paragraph)
%\begin{itemize}
%  \item Hexiang work: URU vs non-URU for seq2seq %(automated metrics).
%  \item Kurt's comparison with FlickR retrieval 
%\end{itemize}
%}

%To the purpose of comparing the quality of ResNeXt-IG-3.5B feature~\cite{uru} and ResNet feature (152-layers trained on ImageNet~\cite{imagenet}), we first report our caption generation models' performance on COCO caption dataset~\cite{lin2014microsoft}. We evaluate BLEU~\cite{papineni2002bleu}, ROUGE-L~\cite{lin2004rouge}, CIDEr~\cite{vedantam2015cider} and SPICE~\cite{anderson2016spice} and compare model's performances to state-of-the-art models under ~\cite{karpathy2015deep}'s setting. 

\begin{table*}[h]
\small
\begin{center}
\begin{tabular}{|cc|ccccc|}
\hline
Method & Image Encoder & BLEU1 &  BLEU4 &  ROUGE-L  &  CIDEr &  SPICE\\
\hline
Human & - & 66.3 & 21.7 & 48.4 & 85.4 & 19.8 \\ \hline
Adaptive~\cite{lu2017knowing} & ResNet & 74.2 & 32.5 & - & 108.5 & 19.5 \\
Att2in~\cite{rennie2017self}  & ResNet & - & 33.3 & 55.3 & 111.4 & - \\
NBT~\cite{lu2018neural}  & ResNet & 75.5 & 34.7 & - & 107.2 & 20.1\\ 
\UpDown~\cite{anderson2017bottom} & ResNet FRCNN & \textbf{79.8} & 36.3 & 56.9 & 120.1 & \textbf{21.4} \\ \hline
\ShowTell (Our)    & ResNet152 & 75.2	& 31.5 & 54.2 & 103.9 & 18.4 \\
\ShowAttTell (Our) & ResNet152 & 76.5	& 32.4 & 55.1 & 109.7 & 19.2 \\
\UpDown (Our) & ResNet152 & 77.0	& 33.9 & 55.6 & 112.7 & 19.6 \\ \hline
\ShowTell  (Our) & ResNeXt-IG-3.5B & 78.2 & 35.0 & 56.6 & 119.9 & 20.8 \\
\ShowAttTell (Our) & ResNeXt-IG-3.5B & 78.8 & 35.6 & 57.1 & 121.8 & 20.6 \\
\UpDown (Our) & ResNeXt-IG-3.5B & 79.3 & \textbf{36.4} & \textbf{57.5} & \textbf{124.0} & 21.2 \\ 
\hline
\end{tabular}
\caption{Generative model performance on COCO caption using the test split of ~\cite{karpathy2015deep}
\label{autometric_coco_caption}
}
\end{center}

\end{table*}

\begin{table*}[t]
\begin{center}
\setlength{\tabcolsep}{0.6em}
\small
\begin{tabular}{|lc|ccc|ccc|}

\hline

\multicolumn{1}{|c}{ Model} & { Text Pre-} & \multicolumn{3}{c}{ Flickr30k} & \multicolumn{3}{c|}{ COCO} \\ 
\multicolumn{1}{|c}{ }  &  { training}
 & R@1 & R@5 & R@10 & R@1 & R@5 & R@10\\
\hline
UVS \cite{DBLP:journals/corr/KirosSZ14}    &       -         & 23.0 & 50.7 & 62.9 & 43.4 & 75.7 & 85.8\\
Embedding Net \cite{8268651}          &             -        & 40.7 & 69.7 & 79.2 & 50.4 & 79.3 & 69.4\\
sm-LSTM \cite{DBLP:journals/corr/HuangWW16}    &      -      & 42.5 & 71.9 & 81.5 & 53.2 & 83.1 & 91.5\\
VSE++ (ResNet, FT) \cite{DBLP:journals/corr/FaghriFKF17}& -  & 52.9 & 80.5 & 87.2 & 64.6 & {90.0} & 95.7\\
GXN (i2t+t2i) \cite{DBLP:journals/corr/abs-1711-06420}  & -  & 56.8 & -    & 89.6 & \textbf{68.5} & -    & \textbf{97.9} \\ 
\hline
{\em \ourmodel~model variants:} & & & & & & &  \\
~Transformer, ResNet152           &      Full           & 10.3 & 27.3 & 38.8 & 21.7 & 45.6 & 58.9\\
~Bag of words, ResNeXt-IG-3.5B      &  None      & 50.0 & 81.1 & 90.0 & 51.6 & 85.3 & 93.4\\
~Transformer, ResNeXt-IG-3.5B     &     None     & 55.6 & 83.2 & 90.5 & 64.0 & 90.6 & 96.3\\
~Bag of words, ResNeXt-IG-3.5B     &   Word             & 58.6 & 87.2 & 92.9 & 54.7 & 87.1 & 94.5\\
~Transformer,  ResNeXt-IG-3.5B &       Word              & \textbf{68.4} & \textbf{90.6} & \textbf{95.3} & 67.3 & {\textbf{91.7}} & 96.5\\
\hline
\end{tabular}
\caption{Retrieval model performance on Flickr30k and COCO caption using the splits of ~\cite{karpathy2015deep}. COCO caption performance is measured on the 1k image test split.
\vspace{-4mm}
\label{retrieval_model_performance_traditional}}
\end{center}
%
%\end{table*}
%
%\begin{table*}[h]
%

\end{table*}

\begin{table*}
\begin{center}
\small
\begin{tabular}{|ccc|ccccc|}
\hline

Method & Image Encoder & { Personality} &  BLEU1 &   BLEU4 &  ROUGE-L & CIDEr &  SPICE\\
\hline
Human Baseline  & - & Yes & 30.1 & 2.8	& 20.1 & 10.8 & 5.1 \\ 
\hline
\ShowTell       & ResNet152 & No & 35.6	& 3.6 & 21.5 & 6.0 & 2.2 \\
\ShowAttTell    & ResNet152 & No & 37.8	& 4.5 & 23.2 & 9.3 & 3.3 \\
\UpDown         & ResNet152 & No & 36.8	& 4.1 & 22.8 & 8.8 & 3.2 \\ 
\hline
\ShowTell      & ResNet152 & Yes & 39.7	& 7.2 & 25.0 & 9.6 & 1.8  \\
\ShowAttTell   & ResNet152 & Yes & 42.7	& 7.2 & 26.8 & 12.4 & 3.8 \\
\UpDown        & ResNet152 & Yes & 43.9	& \bf 8.0 & 27.3 & 13.6 & 3.9 \\ 
\hline
\ShowTell      & ResNeXt-IG-3.5B & No & 36.5 & 4.5 & 22.2 & 7.8	& 2.4 \\
\ShowAttTell   & ResNeXt-IG-3.5B & No & 38.5 & 4.9 & 23.5 & 11.4 & 4.0 \\
\UpDown        & ResNeXt-IG-3.5B & No & 38.9 & 4.8 & 23.5 & 12.0 & 4.1\\
\hline
\ShowTell      & ResNeXt-IG-3.5B & Yes & 38.4 & 7.3 & 24.3 & 9.6 & 1.6 \\
\ShowAttTell   & ResNeXt-IG-3.5B & Yes & 43.3 & 7.1	& 27.0 & 12.6 & 3.6 \\
\UpDown        & ResNeXt-IG-3.5B & Yes & \bf 44.0 & \bf 8.0 & \bf 27.4 & \bf 16.5 & \bf 5.2 \\

%\ShowTell$^\dagger$  & ResNeXt-IG-3.5B & Yes & 15.6 & 1.4	& 14.7 & 17.0 & 3.0 \\
%\ShowAttTell$^\dagger$  & ResNeXt-IG-3.5B & Yes & 15.0 & 1.5 & 14.9 & 18.5 & 4.8 \\
%\UpDown$^\dagger$ & ResNeXt-IG-3.5B & Yes & \textbf{16.4} & \textbf{1.6} & \textbf{15.5} & 21.5 & \textbf{7.5} \\
\hline
\end{tabular}
\caption{Generative model caption performance on the \textsc{\personacaptions} test set. %For our models, $^\dagger$ indicates a model with pretrained word embeddings.}
\label{autometric_persona_caption}
}
\end{center}
\vspace*{-4mm}
\begin{center}
\small
\begin{tabular}{|l|c|c|c|c|c|}
\hline
\multicolumn{1}{|l|}{ Text Encoder} &  Pre-training &\multicolumn{1}{c}{Image Encoder} &\multicolumn{1}{|c}{Personality Encoder} &\multicolumn{1}{|c|}{R@1}\\
%& & & Test \\
\hline
% Transformer & None & None & Yes & 14.5\\
Transformer & None & None & Yes & 20.0\\
% Transformer & Full & None & Yes & 18.1 & 25.8\\
Transformer & Full & None & Yes & 25.8\\
\hline
% Transformer  & Full  &	ResNet152                                & { No}   &	16.6 &  18.7\\
Transformer  & Full  &	ResNet152                                & { No}   & 18.7\\
% Bag of Words  &  None  &	ResNet152                                & Yes  &	24.2 & 35.4\\
Bag of Words  &  None  &	ResNet152                                & Yes  &	35.4\\
% Transformer  & None &	ResNet152                                & Yes  &	26.8 & 40.6 \\
Bag of Words      &   Word  &	ResNet152                                & Yes  & 40.5\\
Transformer  & None &	ResNet152                                & Yes  &	40.6 \\
% Bag of Words      &   Word  &	ResNet152                                & Yes  &	28.5 & 40.5\\
% Transformer   &  Full &	ResNet152                                & Yes  &	34.4 & 51.7\\
Transformer   &  Full &	ResNet152                                & Yes  & 51.7\\
\hline
% shorthand ResNeXt-101 32$\times$48d IG-3.5B-17k -> ResNeXt-IG-3.5B
% Transformer  &   Full &	ResNeXt-IG-3.5B    & { No}   &	38.5 & 53.9\\
% Bag of Words  & None &	ResNeXt-IG-3.5B    & Yes  &	38.6 & 58.6 \\
% Transformer  & None &	ResNeXt-IG-3.5B    & Yes  &	42.9 & 65.9\\
% Bag of Words & Word  &	ResNeXt-IG-3.5B    & Yes  &	45.7 & 66.2\\
% Transformer	   & Full   & ResNeXt-IG-3.5B    & Yes  &	\textbf{53.5} & \textbf{77.5}\\
Transformer  &   Full &	ResNeXt-IG-3.5B    & { No}   &	53.9\\
Bag of Words  & None &	ResNeXt-IG-3.5B    & Yes  &	58.6 \\
Transformer  & None &	ResNeXt-IG-3.5B    & Yes  &	65.9\\
Bag of Words & Word  &	ResNeXt-IG-3.5B    & Yes  &	66.2\\
Transformer	   & Full   & ResNeXt-IG-3.5B    & Yes  & \textbf{77.5}\\
\hline 
\end{tabular}
\caption{Results for \ourmodel~retrieval variants on the \textsc{\personacaptions} test set. 
\label{table:personachat-retrieval-autometrics}
\vspace{-4mm}
}
\end{center}
\end{table*}

\begin{table*}[t] %!htbp]

\setlength{\tabcolsep}{0.2em}
\center
\begin{tabular}{|c|c|l| }
\hline
\small{Image} & \small{Personality} & \small{Generated comment}  \\  \hline
%Image 15297441
\multirow{ 5}{*}{\includegraphics[height=13ex, width=18.4ex]{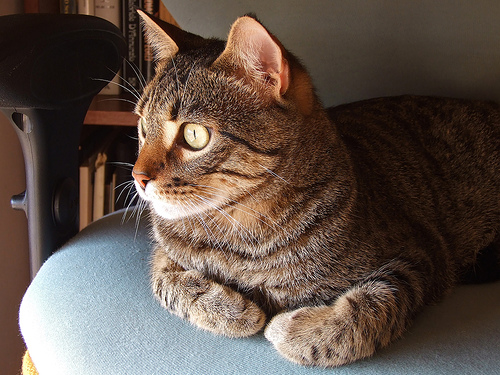}} 
& \small{Anxious} & \small{I love cats but i always get so scared that they will scratch me.}  \rule{0pt}{2.7ex}\\
& \small{Happy} & \small{That cat looks SO happy to be outside.}  \\
& \small{Vague} & \small{That's a nice cat. Or is it a lion?}  \\
& \small{Dramatic} & \small{That cat looks so angry; it might claw your eyes out!}  \\
& \small{Charming} & \small{Awww, sweet kitty. You are so handsome!}  \\ 
\hline
% %Image 79138192
\multirow{ 5}{*}{\includegraphics[height=13ex, width=18.4ex]{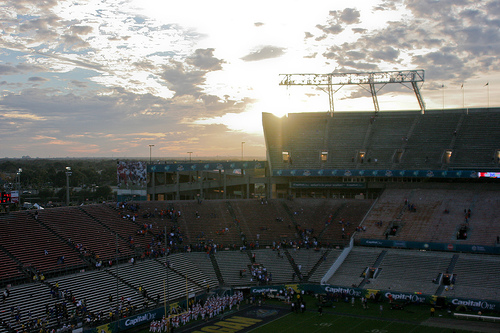}} &
 \small{Sentimental} & \small{The arena reminded me of my childhood.}  \rule{0pt}{2.7ex}\\
& \small{Argumentative} & \small{I dislike the way the arena has been arranged}  \\
& \small{Cultured} & \small{The length of this stadium coincides rather lovely with the width. }  \\
& \small{Sweet} & \small{It was such a nice day at the game. These fans are the best.}  \\
& \small{Romantic} & \small{Basking at the game with my love}  \\
\hline
%Image 53347979
\multirow{ 5}{*}{\includegraphics[height=13ex, width=18.4ex]{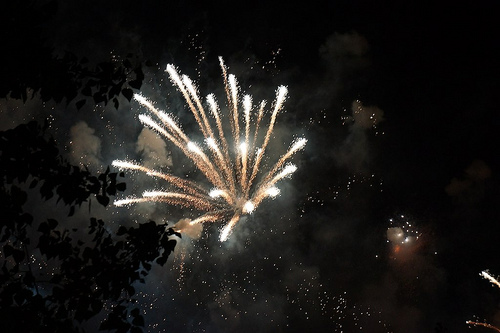}} & \small{Skeptical} & \small{So many fireworks, there is no  way they set them all off at one}  \rule{0pt}{2.7ex}\\
& \small{High-spirited} & \small{Those are the most beautiful fireworks I have ever seen!}  \\
& \small{Cultured} & \small{Fireworks have been used in our celebrations for centuries.}  \\
& \small{Arrogant} & \small{fireworks are overrated and loud}  \\
& \small{Humble} & \small{I'm so grateful for whoever invented fireworks!}  \\
\hline
%Image 67514320
\multirow{ 5}{*}{\includegraphics[height=13ex, width=18.4ex]{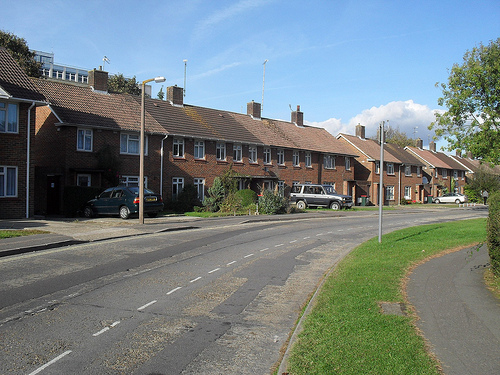}} & \small{Romantic} & \small{A charming home that will call you back to days gone by.}  \rule{0pt}{2.7ex}\\
& \small{Anxious} & \small{This house and this street just makes me feel uneasy.}  \\
& \small{Creative} & \small{I could write a novel about this beautiful old home!}  \\
& \small{Sweet} & \small{What a cute little neighborhood!}  \\
& \small{Money-minded} & \small{Call APR now to get your house renovated!}  \\
\hline
\end{tabular}
\caption{Predictions from our best \ourmodel\xspace model on the \textsc{\personacaptions} valid set. 
% The images come from the validation set, but the comments come from the train set. For aesthetic considerations, we only kept comments of less than 80 characters.
\label{table:model-predictions}
}
\end{table*}

\paragraph{Retrieval Models}

We compare our retrieval architecture, \ourmodel, to existing models reported in the literature on the COCO caption and Flickr30k tasks.
We evaluate retrieval metrics R@1, R@5, R@10, and compare our model performance to state-of-the-art models under the setting of (\cite{karpathy2015deep}).
The results are given in Table
\ref{retrieval_model_performance_traditional} (for more details, see Tables \ref{table:coco_retrieval} and \ref{table:flick30k_retrieval} in the appendix for COCO and Flickr30k, respectively). 
For our model, we see large improvements using ResNeXt-IG-3.5B compared to Resnet152, and stronger performance with a Transformer-based text encoding compared to a bag-of-words encoding.
Pretraining the text encoder also helps substantially (see Appendix \ref{sec:pretraining} for more analysis of pretraining  our systems). 
Our best models are competitive on COCO and are state-of-the-art on Flickr30k by a large margin (68.4 R@1 for our model vs. 56.8 R@1 for the previous state-of-the-art).

\subsection{Automatic evaluations on \personacaptions}
\label{sec:auto_eval_persona}

\paragraph{Generative models} We first train the aforementioned caption generation models without using the personality traits. This setting is similar to standard image captioning, and Table~\ref{autometric_persona_caption} shows that the three caption generation models that we considered are ranked in the same order, with the  \UpDown model %of~\cite{anderson2017bottom} 
being the most effective.
The best results are again obtained using the ResNeXt-IG-3.5B features.
%The best results are obtained using the image featurizer pretrained on Instagram and reach 18.0 CIDEr score.
Adding the embedding of the personality trait allows our best model to reach a CIDEr score of 16.5,
showing the importance of modeling personality in our new task.
%This result is consistent with intuition since we introduce more relevant information. 

Note that all scores are lower than for the COCO captioning task. Indeed standard image captioning tries to produce text descriptions that are semantically equivalent to the image, whereas \textsc{\personacaptions} captures how a human 
responds to a given image when speaking to another human when both can see the image -- which is rarely to simply state its contents.
\textsc{\personacaptions} has
intrinsically more diverse outputs, similar to results found in other human communication tasks \cite{liu2016not}.
%In that sense traditional image captioning is closer to machine translation, whereas our task is closer to a dialogue task, which are known to behave very differently in terms of automatic metrics, e.g. the diversity of language becomes large, see \cite{liu2016not}. %In our task, similar to the situation in dialogue research the diversity of language becomes large, see \cite{liu2016not}.
Besides, as in COCO \cite{chen2015microsoft}, measures like BLEU do not correlate well with human judgements (see top row in Tables \ref{autometric_coco_caption} and \ref{autometric_persona_caption})
hence 
%For that reason 
we perform human evaluation of our models in Section \ref{sec:human_eval}.

\paragraph{Retrieval models} Similarly we compare the effect of various configurations of our retrieval model, \ourmodel. The models are
evaluated in terms of R@1, where for each sample there are 500 candidates to rank:  495 randomly chosen candidates from the test set plus the true labels.

Table \ref{table:personachat-retrieval-autometrics} shows the scores obtained on the test set of \textsc{\personacaptions}.  %, and differ for each sample (but  the same candidates are considered for all models).
%In both generative and retrieval setting, we show that conditioning on persona is possible and significantly improves the result over all considered metrics. In addition, we find that the image features from \citeauthor{uru} also significantly improves the scores in all cases.
Again, the impact of using the image encoder trained on billions of images is considerable, we obtain 77.5\% for our best  ResNeXt-IG-3.5B model,
and 51.7\% for our best Resnet152 model.
 %This underlines the importance of the image analysis part in image captioning tasks. 
Conditioning on the personality traits is also very important 
(77.5\%  vs. 53.9\% R@1 for the best variants with and without conditioning). Transformer text encoders also outperform bag-of-word embeddings encoders, where pretraining for either type of encoder helps.
For Transformers pretraining the whole network performed better than just pretraining the word embeddings, see Appendix \ref{sec:pretraining}.

Example predictions of our best model, \ourmodel~(ResNeXt-IG-3.5B), are given in Table
\ref{table:model-predictions}.
 
% As element of comparison, the same model without the personality information still scores 38.5\% R@1. Intuitively, the tone of the caption seems a crucial information when trying to retrieve the right caption associated with a picture, but its impact is actually dwarfed by the importance of the image processing. This leads to recommend carefulness when comparing two image captioning systems with different image featurization.

\subsection{Human Evaluation on Personality-Captions} % \textsc{\personacaptions}}
 \label{sec:human_eval}
 
% \sh{
% \begin{itemize}%     \item Much more engaging 
% 	\item Positive has a bigger gap, but negative still better
% 	\item Personality vs. Engaging
% 	\item Turkers prefer (marginally) captions produced with a positive personality vs. those produced that are just “engaging”
% 	\item Turkers prefer engaging (or positive) personas to negative personas
% 	\item People are more diverse when you give them a personality: requires classifiers (I have to say I didn't understand that line)
% 	\item Annotators: they liked the engaging one
% 		            Writing engaging captions is more enjoyable than not
% \end{itemize}
% }
% \sh{
% \begin{itemize}
%   \item retrieval model vs seq2seq model vs turkers, with persona imposed or not
%   \item Engaging caption vs our best model 
% \end{itemize}
% We show that such captions are 
% (i) far more engaging to humans
% than traditional ones, and (ii) compared to simply 
% asking humans
% for an engaging caption {\em without conditioning} on a given trait our captions are more controllable, diverse and give higher engagement scores for certain traits. 

The goal of \textsc{\personacaptions} is to be engaging  by emulating human personality traits.
We thus test our task and models in a set of human evaluation studies. 

%the engagingness of \textsc{\personacaptions} captions in an effort to show the significance of conditioning on personalities (for both human annotators and in image captioning models). We measure through human evaluation whether such captions are more engaging than traditional ones. We also compare them to another set of captions generated by humans with the constraint of being engaging, but without personalities.

\paragraph{Engagingness Evaluation Setup} 
%For a given evaluation image, we 
%We conduct these experiments in two parts: first, we collected captions for images; then, we measured relative engagingness between pairs of captions for the each image. 
Using 500 random images from the YFCC-100M dataset that are not present in \textsc{\personacaptions}, we obtain captions for them using a variety of methods, as outlined  below, including both human authored captions and model predicted captions. 
Using a large separate set of human crowdworkers, comparisons are then done pairwise: we show each image, with two captions to compare, to five separate annotators and give them the instruction: 
``The goal of this task is to pick which comment is the most engaging (interesting, captivating, attention-grabbing)''. This results in 2500 trials in total for each pairwise comparison test. 
For experiments where both captions are conditioned on a personality, we show the annotator the personality; otherwise, the personality is hidden. We then report the percentage of the time one method is chosen over the other. 
% The results are summarized in Table \ref{humanevals1}.
The results are given in Table \ref{fig:humanevals1}.

\begin{table*}[!htbp]
\begin{center}
\small
\begin{tabular}{|c|C{1cm}|C{1cm}|c|}
\hline
Type of caption A & \multicolumn{2}{c|}{{\textsc{Win Percentage}}} & Type of caption B \\
\hline
Human personality captions &   \bf{64.5}  &  35.5  & Human traditional captions \\
\hline
\hline
Human personality captions &  \bf{50.5} & 49.5 & TransResNet (ResNeXt-IG-3.5B) \\
\hline
Human personality captions &  \bf{59.1} & 40.9 & TransResNet (ResNet-152) \\
\hline
Human personality captions &  \bf{79.3} & 20.7 & UpDown (ResNeXt-IG-3.5B) \\
\hline
\hline
TransResNet (ResNeXt-IG-3.5B) &  \bf{55.2} & 44.8 & TransResNet (ResNet-152) \\
\hline
TransResNet (ResNeXt-IG-3.5B)$^*$ &  \bf{80.1} & 19.9 & UpDown (ResNeXt-IG-3.5B) \\
\hline
\end{tabular}
\caption{Human evaluations on \textsc{\personacaptions}. Engagingness win rates of various pairwise comparisons:
human annotations of \textsc{\personacaptions} vs. traditional captions,  vs. \textsc{\personacaptions} model variants, and models compared against each other. Our best model {TransResNet (ResNeXt-IG-3.5B)} is close to human performance.
% \ourmodel~vs. other model variants.
%compared to human traditional captions, human engaging captions (without explicit personality conditioning), and two versions of our model.
\label{fig:humanevals1}
}
\end{center}

\end{table*}

\paragraph{Traditional Human Captions} 
We also collected traditional neutral (COCO-like) captions for our 500 test images.
Specifically, the instructions were ``You will be shown an image, for which you will provide a caption'' with the example ``E.g, if you are shown an image of a snow-covered tree in a park, you could write {\em A tree in a park, covered with snow}''.
We then compared human authored \textsc{\personacaptions} captions to these neutral captions.
%human authored traditional neutral (COCO-like) captions.
Captions conditioned on a personality were found to be significantly more engaging than the neutral captions, with a win rate of 64.5\%, which is statistically significant using a binomial
two-tailed test ($p < .001$).

%We performed two engagingness evaluations for our \textsc{\personacaptions} captions compared to captions produced by humans via other methods. %For these evaluations, we asked separate sets of human annotators (who had never participated in the original dataset collection) to write new captions on the 500 selected images. 

%In the first evaluation, we asked annotators to write a caption for the image. We found that annotators believed that the captions conditioned on a personality were significantly more engaging than those that were captions of the image. 

%However, when splitting the captions on the personality trait type in \textsc{\personacaptions} captions, we found that the human annotators preferred captions produced with a positive trait when compared to ``engaging" captions unconditioned on a personality. %We believe that annotators might not find the captions conditioned on neutral and negative personalities to be as engaging without any context of the personality.
%\todo{link to appendix for diversity analysis}
%Note that in addition to this finding, we also found our captions to be more diverse, as well as more controllable than an engaging-only setup; see Appendixes \ref{sec:engaging_no_personality} and \ref{sec:diversity}.

\paragraph{Human vs. Model Engagingness} We compare the  best-performing models from Section 
\ref{sec:auto_eval_persona} to human
authored \textsc{\personacaptions} captions. For each test image we condition both human and model on the same (randomly-chosen) personality trait.
%, we generated captions for the 500 images conditioned on the same personality given to the original annotators.
%In the evaluation experiments, annotators were shown the personality used to write the caption, and were asked to choose the more engaging caption of the two candidates. 
Our best \ourmodel~model from Sec. \ref{sec:auto_eval_persona}, using the ResNext-IG-3.5B image features,  almost matched human authors,
with a win rate of $49.5\%$ (difference not significant, $p>0.6$). 
%This is promising as even a small improvement over our models could render superhuman performance on this task.
The same model using
 ResNet152 has a lower win rate of $40.9\%$, showing the importance of strongly performing image features.
 The best generative model we tried, the \UpDown model using ResNext-IG-3.5B image features, performed worse with a win rate of 20.7\%, showing the impact of retrieval for engagement.
 %image features to the original labels, and found a slightly larger gap in engagingness, with the \personacaptions labels winning by a larger margin ($59.1\%$) than they did when compared to our best \ourmodel using the ResNext-IG-3.5B image features.

%We then compared the \personacaptions labels to our best performing generative model, the \UpDown model using ResNext-IG-3.5B image features. In this evaluation, we found that the \personacaptions labels were significantly more engaging than the \UpDown model, as they were chosen $79.3\%$ of the time.

%Finally, we compared the engagingness of our best performing retrieval model, \ourmodel with ResNext-IG-3.5B image features, to two other model variants. 
%We first compared to our best \ourmodel model using ResNet152 image features, and found that the \ourmodel model with the ResNext-IG-3.5B image features outperforms that with ResNet152 image features in terms of engagingness, matching the results using automatic metrics in Section \ref{sec:auto_eval_persona}.

%We then compared to our best generative model, \UpDown with the ResNext-IG-3.5B image features, and found substantial margins of engagingness, with the \ourmodel being chosen over $80\%$ of the time.
\paragraph{Model vs. Model engagingness}  We also compare our models in a pairwise fashion directly, as measured by human annotators. The results given in 
Table \ref{fig:humanevals1} (all statistically significant) show the same trends as we observed before: \ourmodel\xspace with ResNext-IG-3.5B %image features
outperforms the same model with ResNet152 features with a win rate of $55.2\%$, showing the importance of image features. Additionally, \ourmodel~with ResNext-IG-3.5B image features (with no text encoder pretraining, for a fairer comparison, denoted $^*$ in the table) also substantially outperforms \UpDown ResNext-IG-3.5B with a winrate of $80.1\%$.

\paragraph{Human Evaluation of Caption Relevance}

\begin{table}
\begin{center}
\small
\begin{tabular}{|c|c|c|c|}
 \hline
Set of Captions & Fits Personality & Fits Image & Fits both\\
\hline
Human & 83.1\% & 92.8\% & 80.5\% \\
\hline
TransResNet & 87.7\% & 90.2\% & 81.8\% \\
\hline
\end{tabular}
\caption{Human evaluation of caption fit.
\label{table:aggreement}
}
\end{center}
\vspace{-5mm}
\end{table}
\vspace{-1mm}

In addition to our evaluation of engagingness it is important to also check that the produced captions are relevant to the corresponding image and the personality trait. In order to evaluate this we again performed crowd-sourced human evaluation
for the same 500 evaluation images,
%given image, personality trait, caption triples
where we asked annotators if  captions ``fit'' the image and the personality trait. Results are presented in Table \ref{table:aggreement}. Although human captioners are better at fitting the image (92.8\% vs 90.2\%), \ourmodel ~ actually outperforms them at choosing a caption that fits the personality (87.7\% vs 83.1\%). Note that human captioners were not told specifically that their captions should unambiguously fit the personality trait.
%, as the main focus was to provide an engaging personality-based caption. 
Still, our main conclusion is that our model can indeed provide relevant captions.

\section{Conclusion}

In this work we consider models that can simultaneously understand image content and provide engaging captions for humans.
%develop models that can produce image captions engaging to humans, by endowing them with controllable personality traits.
To build strong models, we first leverage the latest advances in image  and sentence encoding to create generative and retrieval models that perform well on standard image captioning tasks. In particular, we
%pre-training on a very large collection of images allows us to 
attain a new state-of-the-art on caption generation on COCO, and introduce a new retrieval architecture, \ourmodel, that yields the highest known R@1 score on the Flickr30k dataset.

To make the models more engaging to humans, we  then condition them  on a set of controllable personality traits.
To that end, we collect a large dataset,  \textsc{\personacaptions} to train such models.
%Using automatic metrics and human evaluations, 
We show that our best system is able to produce  captions that are close to matching human performance in terms of engagement and relevance.
An important open problem that remains is to improve generative models on this task,
which failed to do as well.

\bibliography{cvpr}
\bibliographystyle{ieee}
\newpage\phantom{blabla}
\appendix

\newpage
\section{Impact of Pretrained Word Embeddings and Text Encoders}
\label{sec:pretraining}

\ourmodel~encodes captions using a transformer architecture, which can be pre-trained: 
\begin{itemize}
  \item either by pre-training the word embeddings on a large corpus of text. In this case we used the pre-trained word vector released by FastText \cite{bojanowski2016enriching}
  \item or by pre-training the entire encoder on a similar task, in which case we followed the setting of \cite{training_millions}.
\end{itemize}
Table \ref{table:coco_retrieval}, Table \ref{table:flick30k_retrieval} and Table \ref{table:pcaption_full_retrieval} show several ablation studies showing the importance of this pre-training.

The same word-pretraining can be attempted on generative models as well. Table \ref{table:autometric_persona_caption2} shows that 0.8 BLEU can be gained.

%\section{Retrieval Models on COCO Captions}
%\label{sec:coco_retrieval_appendix}
\begin{table*}[h!]

\begin{center}
\small
\begin{tabular}{|lc|ccc|c|}
\hline
\multicolumn{1}{|c}{\bf Model} & Text Encoder &\multicolumn{4}{c|}{\bf Caption retrieval} \\
                              & Pretraining &  R@1 & R@5 & R@10 & Med Rank\\
\hline
&& \multicolumn{4}{c|}{\bf 1k Images} \\
\hline
m-CNN \cite{7410658}                                      & & 42.8 & - & 84.1 & 2.0 \\
UVS \cite{DBLP:journals/corr/KirosSZ14}                   & & 43.4 & 75.7 & 85.8 & 2.0\\
HM-LSTM \cite{8237470}                                     && 43.9 & - & 87.8 & 2.0 \\
Order Embeddings \cite{DBLP:journals/corr/VendrovKFU15}    && 46.7 & - & 88.9 & 2.0\\
Embedding Net \cite{8268651}                               && 50.4 & 79.3 & 69.4  & -\\
DSPE+Fisher Vector \cite{Wang2016LearningDS}               && 50.1 & - & 89.2  & -\\
sm-LSTM \cite{DBLP:journals/corr/HuangWW16}                && 53.2 & 83.1 & 91.5  & 1.0\\
VSE++ (ResNet, FT) \cite{DBLP:journals/corr/FaghriFKF17}   && 64.6 & 90.0 & 95.7 & 1.0\\
GXN (i2t+t2i) \cite{DBLP:journals/corr/abs-1711-06420}     && 68.5 & - & \textbf{97.9} & 1.0\\ 
\cite{Engilberge_2018_CVPR}      && \textbf{69.8} & \textbf{91.9} & 96.6 & 1.0\\
\hline
Transformer$^\dagger$, Resnet152                            &Word& 21.7 & 45.6 & 58.9 & 7.0\\
Bag of words, ResNeXt-IG-3.5B                               &None& 51.6 & 85.3 & 93.4 & 1.4\\
Bag of words$^\dagger$, ResNeXt-IG-3.5B                     &Word& 54.7 & 87.1 & 94.5 & 1.0\\
Transformer, ResNeXt-IG-3.5B                                &None& 63.4 & 90.6 & 96.3 & 1.0\\
Transformer$^\dagger$, ResNeXt-IG-3.5B                      &Word& 66.6 & 90.6 & 96.3 & 1.0\\
Transformer$^*$, ResNeXt-IG-3.5B                            &Full& 67.3 & 91.7 & 96.5 & 1.0\\
\hline 
& & \multicolumn{4}{c|}{\bf 1k Images} \\
\hline
Order Embeddings \cite{DBLP:journals/corr/VendrovKFU15}    && 23.3 & - & 65.0 & 5.0\\
VSE++ (ResNet, FT) \cite{DBLP:journals/corr/FaghriFKF17}   && 41.3 & 71.1 & 81.2 & 2.0\\
GXN (i2t+t2i) \cite{DBLP:journals/corr/abs-1711-06420}     && 42.0 & - & 84.7 & 2.0\\ 
\hline
Transformer, Resnet152                            &Word& 7.8 & 21.9 & 31.2 & 30.0 \\
Bag of words, ResNeXt-IG-3.5B                               &None& 26.6 & 58.6 & 73.0 & 4.0\\
Bag of words, ResNeXt-IG-3.5B                     &Word& 29.7 & 62.9 & 75.7 & 3.0\\
Transformer, ResNeXt-IG-3.5B                                &None& 38.8 & 71.6 & 82.7 & 2.0\\
Transformer, ResNeXt-IG-3.5B                      &Word& 44 & 73.7 & \textbf{84} & 2.0\\
Transformer, ResNeXt-IG-3.5B                            &Full& \textbf{44.3} & \textbf{74.5} & 83.9 & 2.0\\
\hline
\end{tabular}
\caption{More detailed results for retrieval model performance on COCO Captions %(including rest of validation)
using the splits of ~\cite{karpathy2015deep}. For our \ourmodel~models, we compare two types of pretraining: Full indicates a model with a pretrained text encoder, while Word indicates a model with pretrained word embeddings only.
\label{table:coco_retrieval}}
\end{center}
\end{table*}

%\section{Retrieval Models on Flickr30k}
%\label{sec:flickr30k_retrieval_appendix}
\begin{table*}[htp]

\small
\begin{center}

\begin{tabular}{|lc|ccc|c|}
\hline
\multicolumn{1}{|c}{\bf Model} & Text Encoder & \multicolumn{4}{c|}{\bf Caption retrieval} \\
 & Pretraining & R@1 & R@5 & R@10 & Med Rank\\
\hline
UVS \cite{DBLP:journals/corr/KirosSZ14}                    && 23.0 & 50.7 & 62.9 & 5.0 \\
UVS (Github)                                                && 29.8 & 58.4 & 70.5 & 4.0\\
Embedding Net \cite{8268651}                               && 40.7 & 69.7 & 79.2 & -\\
DAN \cite{DBLP:journals/corr/NamHK16}                      && 41.4 & 73.5 & 82.5 & 2.0\\
sm-LSTM \cite{DBLP:journals/corr/HuangWW16}                && 42.5 & 71.9 & 81.5 & 2.0\\
2WayNet \cite{DBLP:journals/corr/EisenschtatW16}           && 49.8 & 67.5 & - & -\\
VSE++ (ResNet, FT) \cite{DBLP:journals/corr/FaghriFKF17}   && 52.9 & 80.5 & 87.2 & 1.0\\
DAN (ResNet) \cite{DBLP:journals/corr/NamHK16}             && 55.0 & 81.8 & 89.0 & 1.0\\
GXN (i2t+t2i) \cite{DBLP:journals/corr/abs-1711-06420}     && 56.8 & - & 89.6 & 1.0\\ 
\hline
Transformer, Resnet152                            &Word& 10.3 & 27.3 & 38.8 & 19\\
Bag of words, ResNeXt-IG-3.5B                     &None& 50.0 & 81.1 & 90.0 & 1.5\\
Transformer, ResNeXt-IG-3.5B                      &None& 55.6 & 83.2 & 90.5 & 1.0\\
Bag of words, ResNeXt-IG-3.5B                     &Word& 58.6 & 87.2 & 92.9 & 1.0\\
Transformer, ResNeXt-IG-3.5B                      &Full& 62.3 & 88.5 & 94.4 & 1.0\\
Transformer, ResNeXt-IG-3.5B                      &Word& \textbf{68.4} & \textbf{90.6} & \textbf{95.3} & 1.0\\
\hline
\end{tabular}
\end{center}
\caption{Retrieval model performance on Flickr30k using the  splits of ~\cite{karpathy2015deep}. For our models, we compare two types of pretraining: Full indicates a model with a pretrained text encoder, while Word indicates a model with pretrained word embeddings only.
\label{table:flick30k_retrieval}
}
\end{table*}

\begin{table*}[h]
\setlength{\tabcolsep}{0.4em}
\begin{center}
\small
\begin{tabular}{|lcc|ccccc|}
\hline
Method & Image Encoder & {\small Personality} &  BLEU1 &  BLEU4 &  ROUGE-L &  CIDEr &  SPICE\\
\hline
{\em no pretraining:} & & & & & & & \\
\ShowTell      & ResNeXt-IG-3.5B & Yes & 38.4 & 7.3 & 24.3 & 9.6 & 1.6 \\
\ShowAttTell   & ResNeXt-IG-3.5B & Yes & 43.3 & 7.1	& 27.0 & 12.6 & 3.6 \\
\UpDown        & ResNeXt-IG-3.5B & Yes & 44.0 & 8.0 & 27.4 & \bf 16.5 & \bf 5.2 \\\\
\hline
\multicolumn{2}{|l}{\em with word embedding pretraining:} & & & & & & \\
\ShowTell$^\dagger$     & ResNeXt-IG-3.5B & Yes & 40.1 & 7.7 & 25.3 & 11.0 & 2.2 \\
\ShowAttTell$^\dagger$  & ResNeXt-IG-3.5B & Yes & 44.6 & 7.5 & 25.9	& 12.6 & 3.6 \\
\UpDown$^\dagger$       & ResNeXt-IG-3.5B & Yes & \bf 44.8 & \bf 8.1 & \bf 27.7 & 16.3 & \bf 5.2 \\
\hline
\end{tabular}
\caption{Comparing Generative model caption performance on the \textsc{\personacaptions} test set: pretrained word embeddings vs. no pretraining. Pretraining makes a very small impact in this case, unlike in our retrieval models. \label{table:autometric_persona_caption2}}
\end{center}
\end{table*}

%\section{Retrieval Models on \textsc{\personacaptions}}
%\label{sec:personachat_retrieval_appendix}
\begin{table*}[htp]

\begin{center}
\small
\begin{tabular}{|lc|ccc|}
\hline
\multicolumn{2}{|c}{Text Encoder} & & & \\ 
Encoder Type  & Pretraining &  Image Encoder & Personality Encoder & R@1\\
\hline
Transformer  & Full	      & ResNeXt-IG-3.5B    & Yes  &	\textbf{77.5} \\
Transformer & Word &	ResNeXt-IG-3.5B   & Yes  &	71.7 \\
Bag of Words & Word&	ResNeXt-IG-3.5B    & Yes  &	66.2 \\
Transformer   & None       &	ResNeXt-IG-3.5B    & Yes  &	65.9 \\
Bag of Words   & None      &	ResNeXt-IG-3.5B    & Yes  &	58.6\\
Transformer & Full	      &	ResNeXt-IG-3.5B    & No   &	53.9 \\
Transformer & Full	      &	Resnet152                                & Yes  &	51.7\\
Transformer & Word &	Resnet152                                & Yes  &	45.4\\
Transformer    &  None     &	Resnet152                                & Yes  &	40.6\\
Bag of Words & Word&	Resnet152                                & Yes  &	40.5\\
Bag of Words &   None       &	Resnet152                                & Yes  &	35.4\\
Transformer & Full      &	Resnet152                                & No   &	18.7 \\
\hline
\end{tabular}
\end{center}
\caption{Retrieval model performance on \textsc{\personacaptions}. We compare two types of pretraining: Full indicates a model with a pretrained text encoder, while Word indicates a model with pretrained word embeddings only.
\label{table:pcaption_full_retrieval}
}
\end{table*}

% \section{Statistics of \personacaptions}

% \begin{table}[h!]
% \begin{center}
% \small
% \begin{tabular}{|l|ccc|cc|}
% \hline
%  & \multicolumn{3}{|c|}{{\bf \personacaptions}} \\
%  \hline
% Split & train & valid & test\\
% \hline
% Number of Examples &    186,858  &  5,000  & 10,000 \\
% \hline
% Number of Personality Types  &  215 & 215 & 215 \\
% Vocabulary Size & 35559  & 5557 & 8137 \\
% Average Tokens per Caption   & 11.6 & 11.2 &11.4  \\
% \hline
% \end{tabular}
% \caption{\textsc{\personacaptions} dataset statistics.
% \label{table:data_stats}
% }
% \end{center}
% \end{table}

% \begin{figure}[h!]
% \begin{center}
% \includegraphics[width=0.5\textwidth]{Winrate_Models.png}
% \caption{Win rate of \ourmodel~annotations
% compared to other model variants.
% \label{fig:humanevals2}
% }
% \end{center}
% %\end{figure}
% %\begin{figure}[h!]
% \end{figure}

\section{Engaging Captions, with no personality conditioning}
\label{sec:engaging_no_personality}

\paragraph{Engaging-only Captions} 
Instead of asking to author a caption based on a personality trait, we can ask humans to simply write an ``engaging'' caption
instead, providing them with no personality cue.
%We asked annotators to write ``engaging" captions, but did not provide them with personalities. 
We found that human annotators overall preferred unconditioned captions to those conditioned on a personality by a slight margin ($\sim54\%$). To further understand this difference, we split the images into three subsets based on the personality on which the \textsc{\personacaptions} annotator conditioned their caption, i.e. whether the personality was positive, negative, or neutral. We then examined the engagingness rates of images for each of these subsets. In the set where \textsc{\personacaptions} annotators were provided with positive personalities, which totaled 185 out of the 500 images, we found that human annotators preferred the captions conditioned on the personality to those that were not. However, in the other two sets, we found that the unconditioned captions were preferred to the negative or neutral ones. For these two subsets, we believe that, without the context of any personality, annotators may have preferred the inherently more positive caption provided by someone who was asked to be engaging but was not conditioned on a personality.

\begin{table*}[!htbp]
\label{engaging_no_personality_evals}

\begin{center}
\small
\begin{tabular}{|c|C{1cm}|C{1cm}|c|}
\hline
Type of caption A & \multicolumn{2}{c|}{{\textsc{Win Percentage}}} & Type of caption B \\
\hline
Human (all) personality captions &   45.5  &  \bf{54.5}  & Human engaging captions \\
\hline
Human (positive) personality captions  &  \bf{51.2} & 48.8 & Human engaging captions \\
\hline

\end{tabular}
\caption{Pairwise win rates of various approaches, evaluated in terms of engagingness }
\end{center}
\end{table*}
\paragraph{Diversity of captions}
%\label{sec:diversity}

We found that the captions written via our method were not only more engaging for positive personality traits, but also resulted in more diversity in terms of personality traits. To measure this diversity, we constructed a model that predicted the personality of a given comment. The classifier consists in the same Transformer as described in \ref{retrievalmodels}, pre-trained on the same large dialog corpus, followed by a softmax over 215 units. 
%The model is trained using the ADAM optimizer on the train set of \textsc{\personacaptions} until we do not observe improvements on the validation set. 
We then compare the total number of personality types as predicted by the classifier among each type of human-labeled data: ``engaging" captions conditioned on personalities, ``engaging" captions not conditioned on personalities, and traditional image captions. That is, we look at each caption given by the human annotators, assign it a personality via the classifier, and then look at the total set of personalities we have at the end for each set of human-labeled data. For example, out of the 500 human-generated traditional captions, the classifier found $63\%$ of all possible positive personalities in this set of captions. As indicated in Table \ref{table:diversity}, the human annotators who were assigned a personality produce more diverse captions, particularly negatively and neutrally conditioned ones, as compared to human annotators who are just told to be ``engaging" or those who are told to write an image caption.

\begin{table*}[!htbp]

\begin{center}
\small
\begin{tabular}{|c|ccc|}
\hline
\multicolumn{1}{|c|}{\bf Annotation Task} &\multicolumn{3}{c|}{\bf Personality Trait Coverage} \\
\hline
& Positive & Neutral & Negative\\
Given Personalities           & 100\%  & 100\% & 99.0\% \\
\hline
Traditional Caption           & 63.0\% & 83.3\% & 47.0\% \\
Engaging, No Conditioning     & 81.5\% & 91.7\% & 71.4\% \\
\textsc{\personacaptions} & 82.7\% & 94.4\% & 87.8\% \\
\hline
\end{tabular}
\end{center}
\caption{Caption diversity in human annotation tasks. \textsc{\personacaptions} provides more diverse personality traits than traditional captions or collecting engaging captions without specifying a personality trait to the annotator, as measured by a personality trait classifier. \label{table:diversity}
}
\end{table*}

\section{Comparing Generative and Retrieval Models on COCO}
\label{asec:compare_coco}

The ultimate test of our generative and retrieval models on \textsc{\personacaptions} is performed using human evaluations. Comparing them using automatic metrics is typically difficult because retrieval methods perform well with ranking metrics they are optimized for and generative models perform well with word overlap metrics they are optimized for, but neither of these necessarily correlate with human judgements, see e.g. \cite{zhang2018personalizing}.

Nevertheless, here we compare our generative and retrieval models directly with automatic metrics on COCO. We computed the BLEU, CIDEr, SPICE, and ROUGE-L scores for our best \ourmodel\xspace model. %We additionally compare the scores of (i) a \ourmodel\xspace trained on COCO that retrieves labels from the COCO train set, and (ii) a \ourmodel\xspace trained on \textsc{\personacaptions} that retrieves labels from the \textsc{\personacaptions} train set. 
The comparison is given in Table \ref{table:ret_vs_gen_bleu}. 

\begin{table*}[ht!]
\begin{center}
\small
\begin{tabular}{|c|c|c|c|c|c|}
\hline
Model &  BLEU1 &  BLEU4 &  ROUGE-L  &  CIDEr &  SPICE\\
\hline
%\ourmodel     & \textsc{\personacaptions} & 20.2 & 1.0 & 16.3 & 7.3 & 4.7 \\ 
%\ourmodel     &  COCO                     & 50.6 & 10.9 & 38.0 & 49.1 & 13.9 \\ \hline
%\ourmodel      & 20.2 & 1.0 & 16.3 & 7.3 & 4.7 \\ 
\ourmodel                       & 50.6 & 10.9 & 38.0 & 49.1 & 13.9 \\ \hline
\ShowTell                       & 78.2 & 35.0 & 56.6 & 119.9 & 20.8 \\
\ShowAttTell                    & 78.8 & 35.6 & 57.1 & 121.8 & 20.6 \\
\UpDown                         & 79.3 & \textbf{36.4} & \textbf{57.5} & \textbf{124.0} & 21.2 \\
\hline
\end{tabular}
\caption{Generative and retrieval model performance on COCO caption using the test split of ~\cite{karpathy2015deep}. All models use ResNeXt-IG-3.5B image features.
\label{table:ret_vs_gen_bleu}}
\end{center}
\end{table*}

\begin{figure*}[]
\begin{center}
\includegraphics[trim={0 1mm 0 2mm}, clip, width=0.92\textwidth]{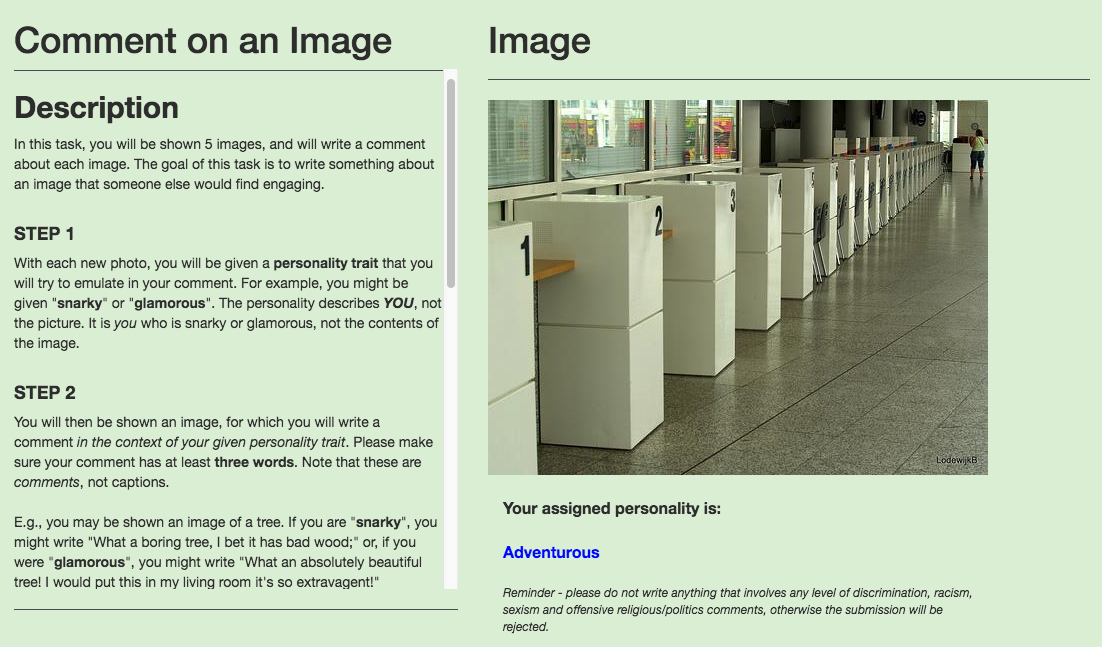}
\caption{Instructions for the annotation task collecting the data for \textsc{\personacaptions}.}
\label{annotation_setup}
\end{center}
\end{figure*}

\begin{table*}[!htbp]

\centering\setlength{\tabcolsep}{0.5em}
\begin{tabular}{p{12em}p{12em}p{12em}}

 & & \\[-0.62em]
\includegraphics[width=12em, height=7.8em]{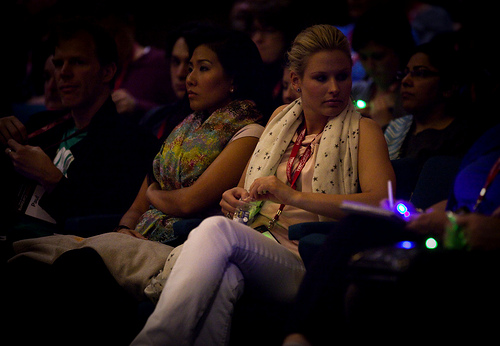} & \includegraphics[width=12em, height=7.8em]{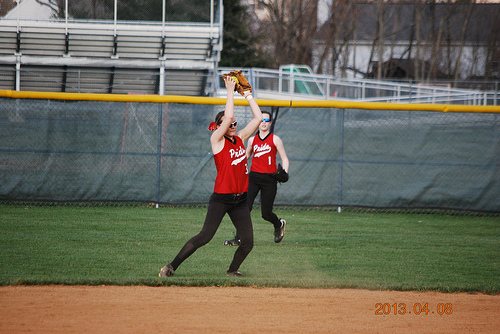} & \includegraphics[width=12em, height=7.8em]{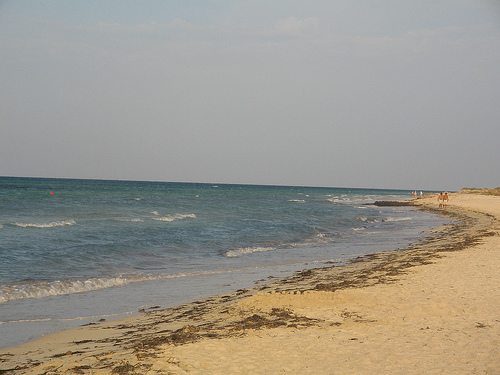}\\
\textit{\small{Sarcastic}} & \textit{\small{Mellow }} & \textit{\small{Zany}} \\ 
\small{Yes please sit by me} & \small{Look at that smooth easy catch of the ball. like ballet.} & \small{I wish I could just run down this shore!} \\

 & & \\[-0.62em]
\includegraphics[width=12em, height=7.8em]{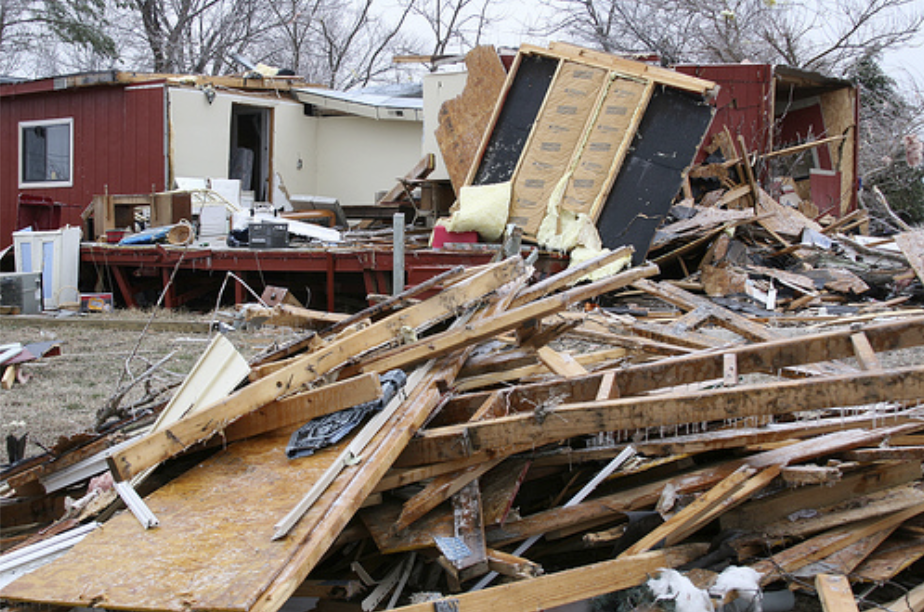} & \includegraphics[width=12em, height=7.8em]{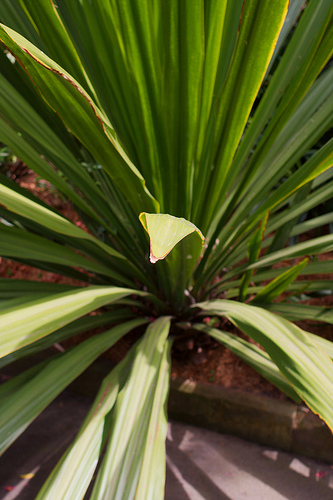} & \includegraphics[width=12em, height=7.8em]{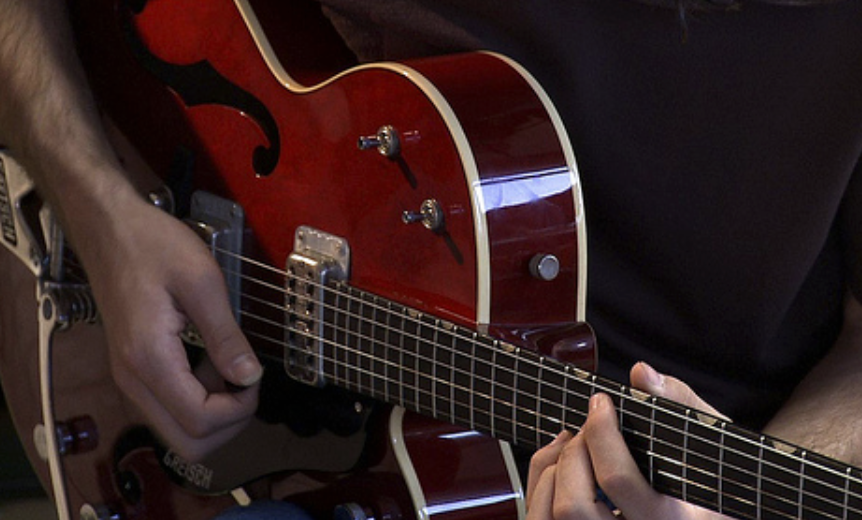}\\
\textit{\small{Contradictory}} & \textit{\small{Contemptible }} & \textit{\small{Energetic}} \\ 
\small{Love what you did with the place!} & \small{I can't believe no one has been taking care of this plant.  Terrible} & \small{About to play the best tune you've ever heard in your life. Get ready!} \\ 

 & & \\[-0.62em]
\includegraphics[width=12em, height=7.8em]{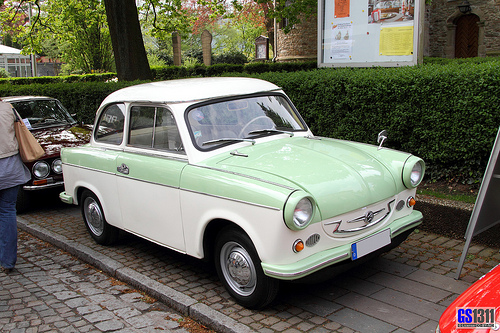} & \includegraphics[width=12em, height=7.8em]{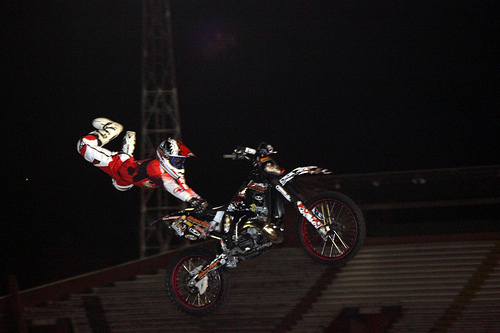} & \includegraphics[width=12em, height=7.8em]{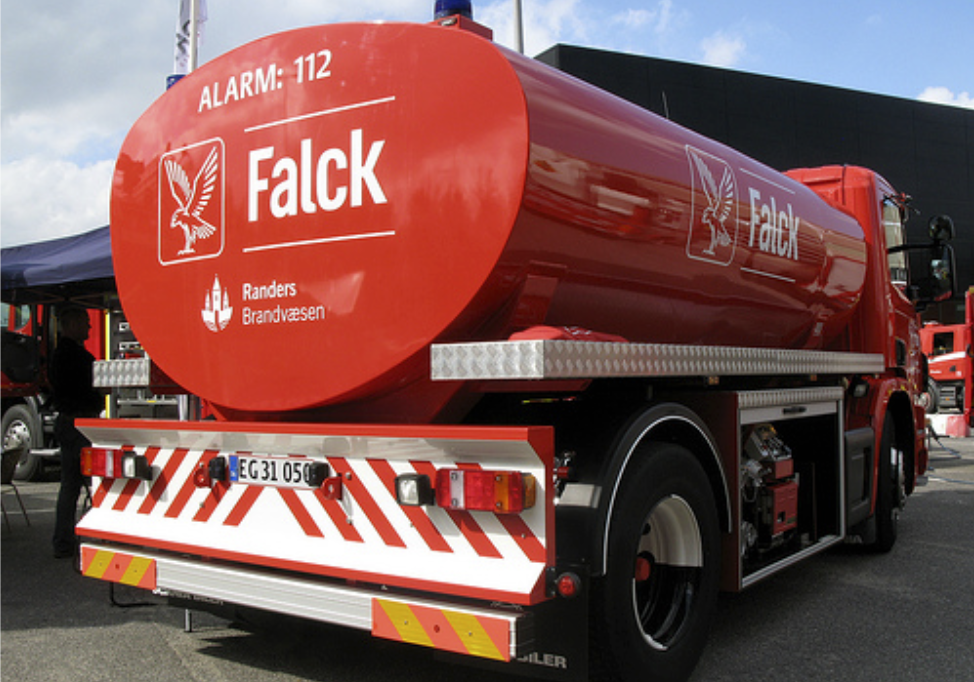}\\
\textit{\small{Kind}} & \textit{\small{Spirited}} & \textit{\small{Creative }} \\ 
\small{they left me a parking spot} & \small{That is one motor cycle enthusiast!!!} & \small{Falck alarm, everyone. Just a Falck alarm.} \\ 

 & & \\[-0.62em]
\includegraphics[width=12em, height=7.8em]{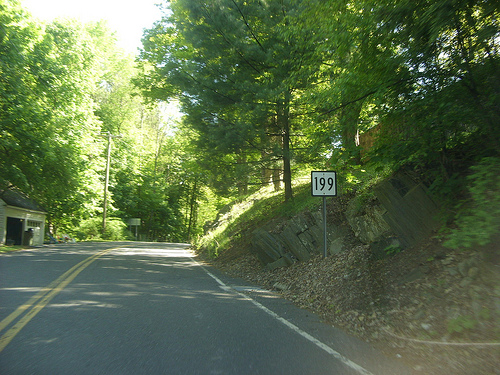} & \includegraphics[width=12em, height=7.8em]{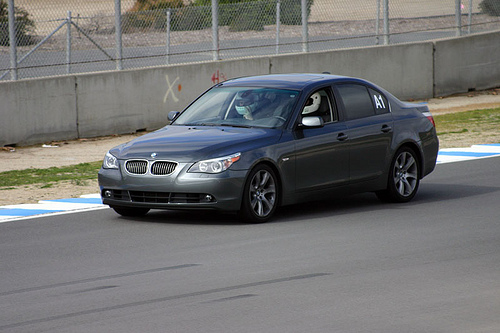} & \includegraphics[width=12em, height=7.8em]{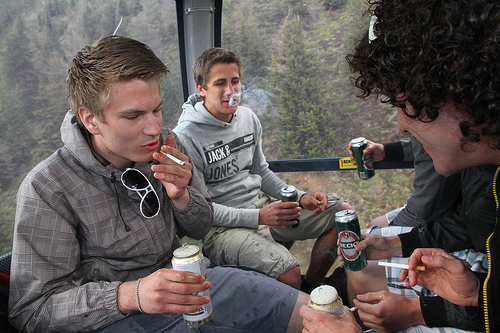}\\
\textit{\small{Crazy}} & \textit{\small{Morbid}} & \textit{\small{Questioning}} \\ 
\small{I drove down this road backwards at 90 miles per hour three times} & \small{I hope this car doesn't get into a wreck.} & \small{Why do people think its cool to smoke cigarettes?} \\

\end{tabular}
\caption{Some samples from \textsc{\personacaptions}. For each sample we asked a person to write a caption that fits both the image and the personality.}

\end{table*}

\begin{table*}[!htbp]
\centering\setlength{\tabcolsep}{0.5em}
\small
\begin{tabular}{p{12em}p{12em}p{12em}}

 & & \\[-0.62em]
\includegraphics[width=12em, height=7.8em]{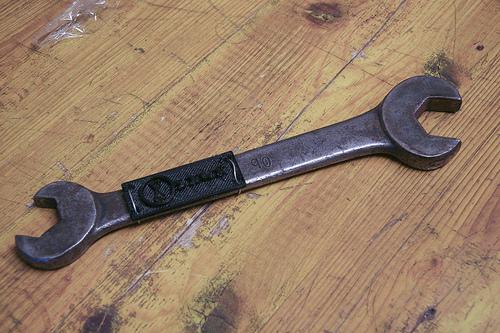} & \includegraphics[width=12em, height=7.8em]{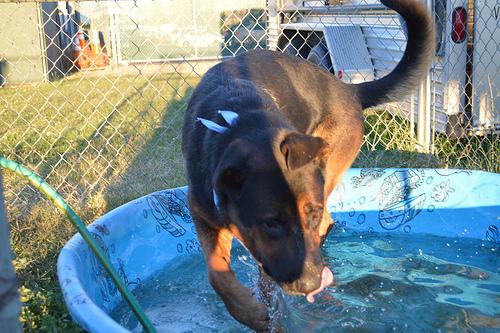} & \includegraphics[width=12em, height=7.8em]{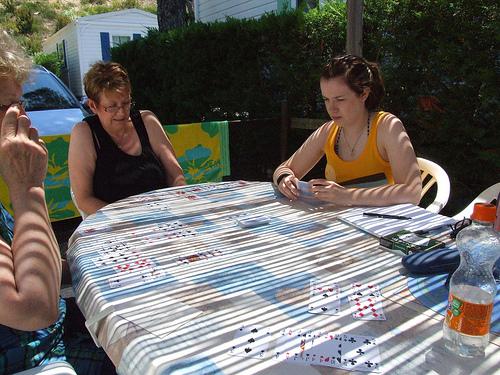}\\
\textit{\small{Old-fashioned}} & \textit{\small{Destructive }} & \textit{\small{Courageous}} \\ 
origin: \ourmodel & origin: \ourmodel & origin: \ourmodel \\
fit: does not fit image & fit: does not fit personality & fit:neither \\
\small{Each of these hammers has a mission.} & \small{that dog is going to drown! someone save it.} & \small{Look at all of those sewing materials! You could create all sorts of art projects with them!} \\

 & & \\[-0.62em]
\includegraphics[width=12em, height=7.8em]{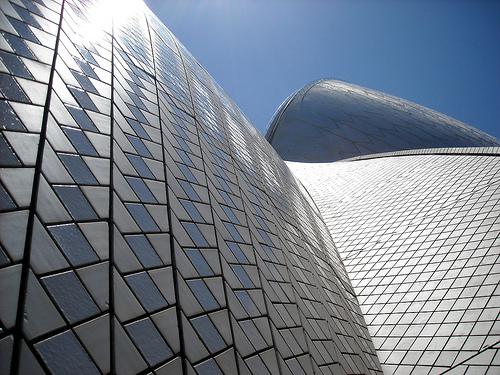} & \includegraphics[width=12em, height=7.8em]{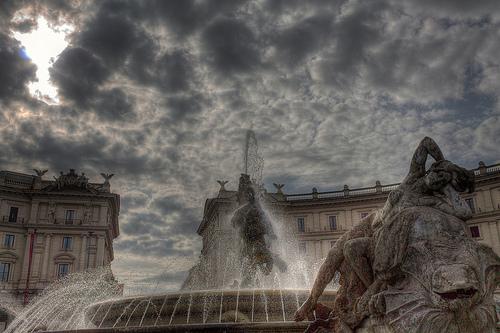} & \includegraphics[width=12em, height=7.8em]{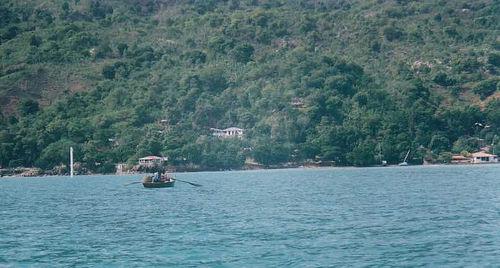}\\
\textit{\small{Meticulous}} & \textit{\small{Sympathetic }} & \textit{\small{Bewildered}} \\ 
origin: human & origin: human & origin: human \\
fit: neither & fit: does not fit personality & fit:neither \\
\small{The desert is so overwhelming and vast I totally want to go exploring again!} & \small{relaxing,calm and authentic} & \small{Graduating school and you finally feel like you're invincible.} \\ 

\end{tabular}
\caption{Some examples of captions that do not fit either the personality or the image, produced by humans and \ourmodel}

\end{table*}

%This table is generated automatically.
\newcolumntype{L}{>{\arraybackslash}p{8cm}}
\begin{table*}[!htbp]
\centering
\begin{adjustbox}{center=10.5cm}\setlength{\tabcolsep}{0.2em}
\begin{tabular}{cclL }

\hline
\small{Image and Pers.} & \small{Use pers.} & \small{Captioning} & \small{Caption}  \\  \hline
%Image 27826118
\\[-1.8ex]
\multirow{6}{*}{\includegraphics[height=12.5ex, width=18ex]{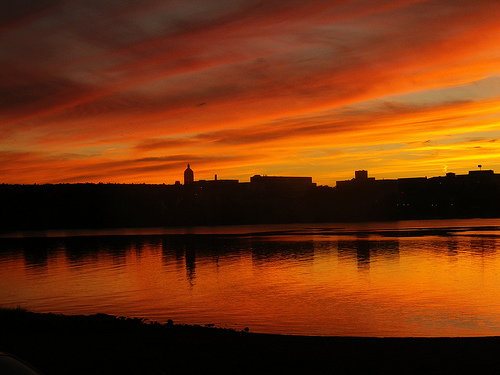}}
\\[-1.1ex]
& No & \small{Standard} & \small{A city on the background, a lake on the front, during a sunset.}  \\
& No & \small{Engaging} & \small{Talk about summer fun! Can I join? :)}  \\
\\
& Yes & \small{Human} & \small{i feel moved by the sunset}  \\
& Yes & \small{\ourmodel} & \small{The water at night is a beautiful sight.}  \\
Spirited & Yes & \small{\UpDown} & \small{This is a beautiful sunset!}  \\[1.4ex] 

\hline
%Image 27826118
\\[-1.8ex]
\multirow{6}{*}{\includegraphics[height=12.5ex, width=18ex]{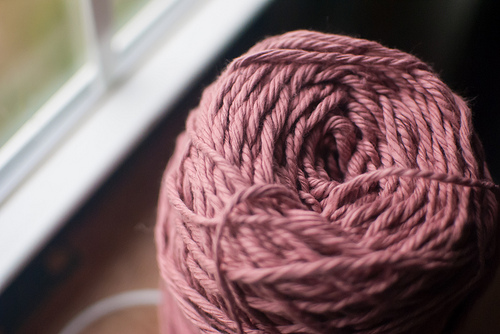}}
\\[-1.1ex]
& No & \small{Standard} & \small{Rose colored soft yarn.}  \\
& No & \small{Engaging} & \small{I really want to untangle that yarn.}  \\
\\
& Yes & \small{Human} & \small{I cannot believe how yummy that looks.}  \\
& Yes & \small{\ourmodel} & \small{What is up with all the knitting on my feed}  \\
Ridiculous & Yes & \small{\UpDown} & \small{I would love to be a of that fruit!}  \\[1.4ex] 

\hline
%Image 27826118
\\[-1.8ex]
\multirow{6}{*}{\includegraphics[height=12.5ex, width=18ex]{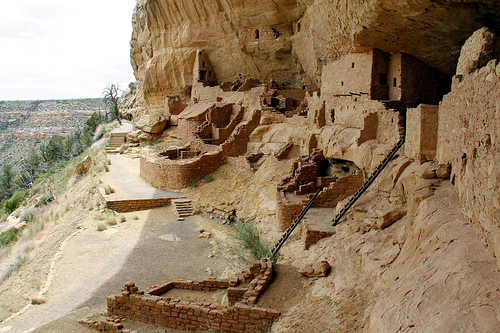}}
\\[-1.1ex]
& No & \small{Standard} & \small{A beautiful mesa town built into the cliffs. }  \\
& No & \small{Engaging} & \small{That is a strange cave}  \\
\\
& Yes & \small{Human} & \small{It must be very dangerous if children play there}  \\
& Yes & \small{\ourmodel} & \small{I hope my kids don't climb on this.}  \\
Maternal & Yes & \small{\UpDown} & \small{I hope this is a beautiful place.}  \\[1.4ex] 

\hline
%Image 27826118
\\[-1.8ex]
\multirow{6}{*}{\includegraphics[height=12.5ex, width=18ex]{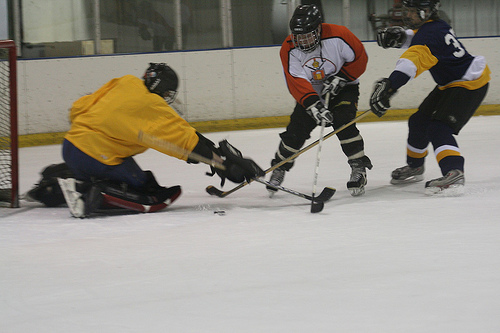}}
\\[-1.1ex]
& No & \small{Standard} & \small{Hockey players competing for control of the hockey puck.}  \\
& No & \small{Engaging} & \small{Great save, goalie!!}  \\
\\
& Yes & \small{Human} & \small{Hockey is a little too barbaric for my taste.}  \\
& Yes & \small{\ourmodel} & \small{Hockey players gracefully skate across the ice.}  \\
Sophisticated & Yes & \small{\UpDown} & \small{This hockey is like they are a great of the game.}  \\[1.4ex]

\hline
%Image 27826118
\\[-1.8ex]
\multirow{6}{*}{\includegraphics[height=12.5ex, width=18ex]{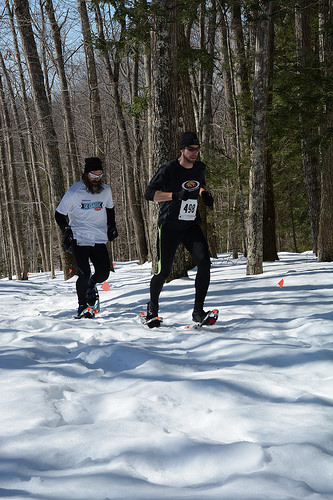}}
\\[-1.1ex]
& No & \small{Standard} & \small{two people walking through a snowy forest.}  \\
& No & \small{Engaging} & \small{Too cold for me.}  \\
\\
& Yes & \small{Human} & \small{I wonder what's at the finish line for these guys?}  \\
& Yes & \small{\ourmodel} & \small{I wonder why they are running.}  \\
Curious & Yes & \small{\UpDown} & \small{I wonder what they are a?}  \\[1.4ex] 

\hline
%Image 27826118
\\[-1.8ex]
\multirow{6}{*}{\includegraphics[height=12.5ex, width=18ex]{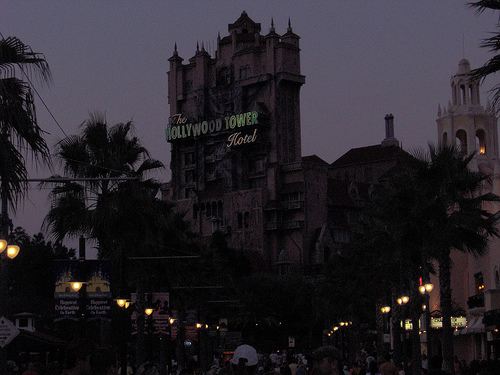}}
\\[-1.1ex]
& No & \small{Standard} & \small{Hollywood Tower at Night}  \\
& No & \small{Engaging} & \small{I went to that theme park, but was too scared to get on that ride!}  \\
\\
& Yes & \small{Human} & \small{I am so excited to be here!}  \\
& Yes & \small{\ourmodel} & \small{I remember going to disney world, it was one of the best trips I've ever done.}  \\
Happy & Yes & \small{\UpDown} & \small{This looks like a beautiful view!}  \\[1.4ex] 

\end{tabular}
\end{adjustbox}
\caption{Example variants of the captions shown to human annotators in the human evaluation tasks in Section \ref{sec:human_eval}. The first two captions are human annotations not conditioned on a personality; the next three are captions conditioned on the listed personality, and are generated via a human annotator, \ourmodel, and \UpDown respectively.} % For aesthetic reasons, we only kept comments under 80 characters long.}
\end{table*}

%This table is generated automatically.
\begin{table*}[!htbp]
\centering
\begin{adjustbox}{center=10.5cm}\setlength{\tabcolsep}{0.2em}
\begin{tabular}{ccl }
\hline
\small{Image} & \small{Personality} & \small{Generated comment}  \\  \hline
%Image 27826118
\\[-1.8ex]
\multirow{ 5}{*}{\includegraphics[height=13.5ex, width=19ex]{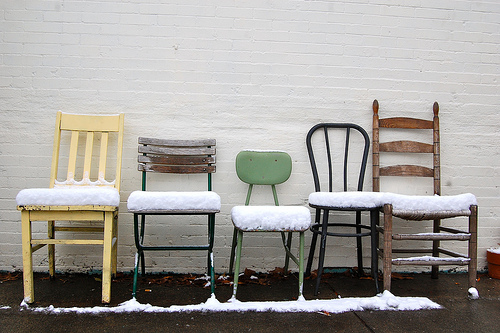}}
\\[-1.5ex]
& \small{Sweet} & \small{I love, love, love these chairs! I want the big one in my house!}  \\
& \small{Vague} & \small{This chair is either covered in snow or the snow is covered in the chair.}  \\
& \small{Cultured} & \small{These chairs remind me of the Swedish interior design revolution of the 70's.}  \\
& \small{Paranoid} & \small{What if someone fell off those chairs.}  \\
& \small{Overimaginative} & \small{Those chairs look like they could be in a doll house.}  \\[1.4ex] 
\hline
%Image 8022120
\\[-1.8ex]
\multirow{ 5}{*}{\includegraphics[height=13.5ex, width=19ex]{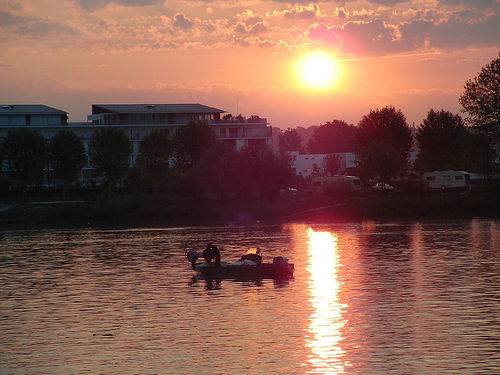}}
\\[-1.5ex]
& \small{Arrogant} & \small{I've seen better sunsets elsewhere.}  \\
& \small{Overimaginative} & \small{that sunset is so orange it could be a fruit}  \\
& \small{Vague} & \small{It's the sunset.}  \\
& \small{Optimistic} & \small{The sunset makes look forward to a happy tomorrow. }  \\
& \small{Charming} & \small{The way the sun is hitting the water makes for a romantic evening.}  \\[1.4ex] 
\hline
%Image 67890626
\\[-1.8ex]
\multirow{ 5}{*}{\includegraphics[height=13.5ex, width=19ex]{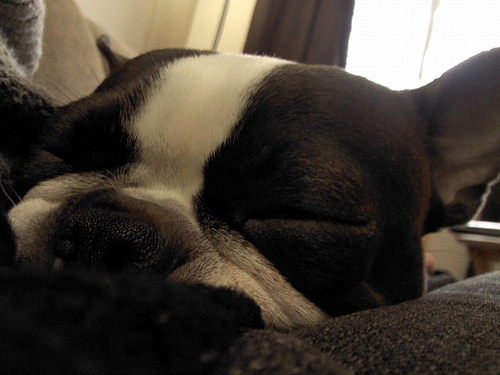}}
\\[-1.5ex]
& \small{Sweet} & \small{What a cute puppy, reminds me of my friends.}  \\
& \small{Skeptical} & \small{I don't think this dog will bite me.}  \\
& \small{Sympathetic} & \small{poor dog! It looks so hungry :c}  \\
& \small{Vague} & \small{it's a dog}  \\
& \small{Wishful} & \small{I wish that I had a dog as cute as him.}  \\[1.4ex] 
\hline
%Image 1803044
\\[-1.8ex]
\multirow{ 5}{*}{\includegraphics[height=13.5ex, width=19ex]{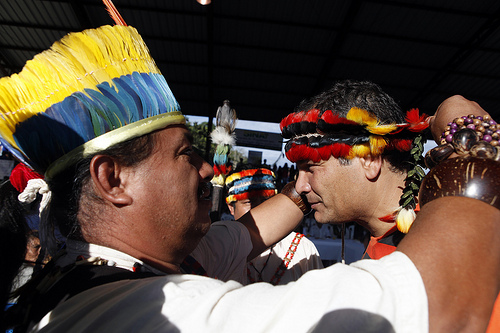}}
\\[-1.5ex]
& \small{Cultured} & \small{I love a cultural celebration.}  \\
& \small{Skeptical} & \small{I'm not sure if these are guys in costumes or time travelers.}  \\
& \small{Sweet} & \small{I love that they are celebrating their traditions and culture.}  \\
& \small{Overimaginative} & \small{They look like they could be dancers in a fantasy movie with dragons!}  \\
& \small{Sympathetic} & \small{I feel sorry for him having to wear that }  \\[1.4ex] 
\hline
%Image 50689277
\\[-1.8ex]
\multirow{ 5}{*}{\includegraphics[height=13.5ex, width=19ex]{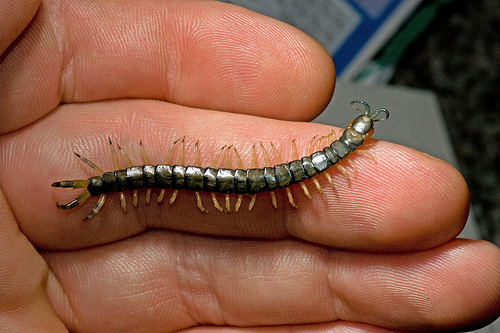}}
\\[-1.5ex]
& \small{Romantic} & \small{If I was an insect, I would definitely make this my mate. }  \\
& \small{Humble} & \small{I am grateful that spiders eat these disgusting bugs.}  \\
& \small{Paranoid} & \small{What is going on? Are these insects dangerous? }  \\
& \small{Creative} & \small{I made something like this from colored toothpicks once}  \\
& \small{Money-minded} & \small{how much are those? those looks expensive}  \\[1.4ex] 
\hline
%Image 20806601
\\[-1.8ex]
\multirow{ 5}{*}{\includegraphics[height=13.5ex, width=19ex]{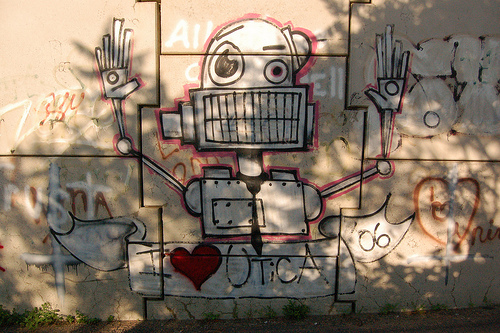}}
\\[-1.5ex]
& \small{Happy} & \small{That is so cool! I I love street art!}  \\
& \small{Optimistic} & \small{The future is bright for people who can dream in artistic ways.}  \\
& \small{Critical} & \small{I do believe this taggers verbage is a tad junvenile}  \\
& \small{Charming} & \small{What a charming wall.}  \\
& \small{Adventurous} & \small{I think I could create art like that, I will go learn and take action.}  \\[1.4ex] 
\hline
%Image 47357199
\\[-1.8ex]
\multirow{ 5}{*}{\includegraphics[height=13.5ex, width=19ex]{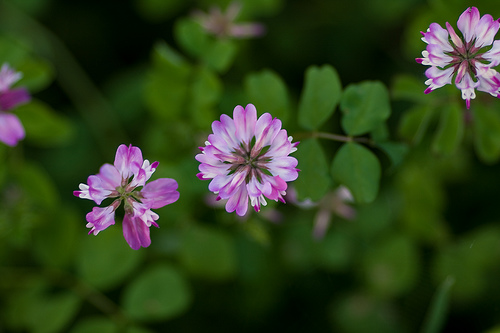}}
\\[-1.5ex]
& \small{Dramatic} & \small{The color of this flower is absolutely astounding.  I can't believe it.}  \\
& \small{Wishful} & \small{I always wish I could grow these types of flowers.}  \\
& \small{Sweet} & \small{Beautiful flowers! I would give them to you. }  \\
& \small{Romantic} & \small{The pink flowers would make a beautiful bouquet for my wife.}  \\
& \small{Happy} & \small{Oh my, what a lovely purple color of nature's new sprouts!}  \\[1.4ex] 
\hline
\end{tabular}
\end{adjustbox}
\caption{More example predictions from our best \textsc{\ourmodel} model on the 
\textsc{\personacaptions} validation set.
}
\end{table*}
%This table is generated automatically.
\begin{table*}[!htbp]
\centering
\begin{adjustbox}{center=10.5cm}\setlength{\tabcolsep}{0.2em}
\begin{tabular}{ccl }
\hline
\small{Image} & \small{Personality} & \small{Generated comment}  \\  \hline
%Image 89307081
\\[-1.8ex]
\multirow{ 5}{*}{\includegraphics[height=13.5ex, width=19ex]{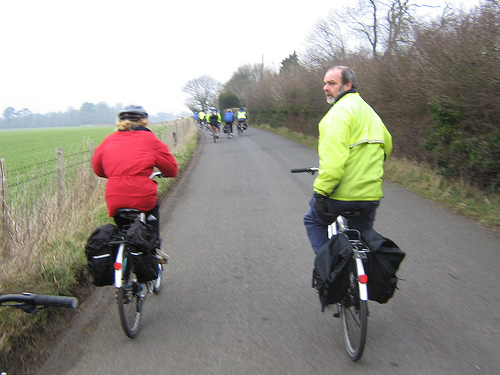}}
\\[-1.5ex]
& \small{Adventurous} & \small{This biking event looks like something that I would try!}  \\
& \small{Vague} & \small{Those people are riding a bike.}  \\
& \small{Charming} & \small{I bet a wonderful couple uses this bike to tour the countryside together.}  \\
& \small{Optimistic} & \small{A hopeful cyclist trying to catch up to the pack }  \\
& \small{Paranoid} & \small{What if all those bikes just tipped over!}  \\[1.4ex] 
\hline
%Image 41593753
\\[-1.8ex]
\multirow{ 5}{*}{\includegraphics[height=13.5ex, width=19ex]{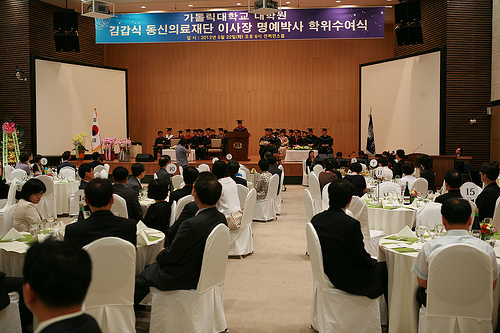}}
\\[-1.5ex]
& \small{Adventurous} & \small{I am so ready for the conference.}  \\
& \small{Cultured} & \small{This conference is one of the most important ones in the country.}  \\
& \small{Vague} & \small{The organization on that table is uncertain.}  \\
& \small{Dramatic} & \small{OMG!! This ceremony is frightening!}  \\
& \small{Sympathetic} & \small{I feel bad for these people being so cramped in this room.}  \\[1.4ex] 
\hline
%Image 91734257
\\[-1.8ex]
\multirow{ 5}{*}{\includegraphics[height=13.5ex, width=19ex]{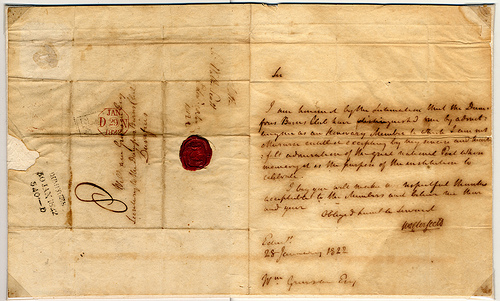}}
\\[-1.5ex]
& \small{Old-fashioned} & \small{Such old fashioned script, a true lost art.}  \\
& \small{Charming} & \small{I could use these to write to my loved ones.}  \\
& \small{Argumentative} & \small{Can you even read this through all the jpeg artifacts?}  \\
& \small{Anxious} & \small{I hope this paper doesnt tear, history will be destroyed.}  \\
& \small{Dramatic} & \small{Some of the most profound things ever written have been on linen. }  \\[1.4ex] 
\hline
% %Image 50313985
% \\[-1.8ex]
% \multirow{ 5}{*}{\includegraphics[height=13.5ex, width=19ex]{images/50313985.jpg}}
% \\[-1.5ex]
% & \small{Cultured} & \small{The moon phase is currently a crescent moon.}  \\
% & \small{Sweet} & \small{The moon is so lovely.}  \\
% & \small{Skeptical} & \small{I am assuming that is the moon.}  \\
% & \small{Paranoid} & \small{The moon will swallow us all.}  \\
% & \small{Overimaginative} & \small{The moon looks like it's a tiny little portal to another dimension.}  \\[1.4ex] 
% \hline
%Image 19088444
\\[-1.8ex]
\multirow{ 5}{*}{\includegraphics[height=13.5ex, width=19ex]{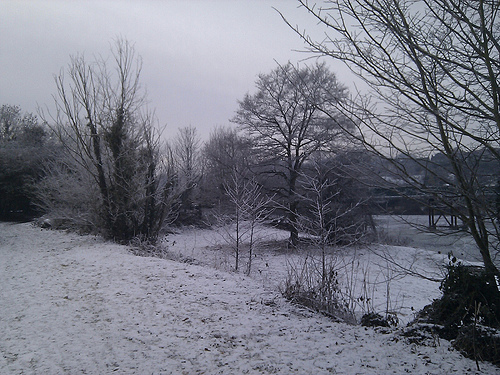}}
\\[-1.5ex]
& \small{Happy} & \small{It finally snowed, it makes me feel awesome}  \\
& \small{Wishful} & \small{I wish there was enough for snow angels.}  \\
& \small{Boyish} & \small{Can I go sledding now?}  \\
& \small{Romantic} & \small{What a beautiful frost! Looks like the perfect place to fall in love!}  \\
& \small{Cultured} & \small{The white of the snow provides a glistening contrast to the dead trees.}  \\[1.4ex] 
\hline
%Image 17646416
\\[-1.8ex]
\multirow{ 5}{*}{\includegraphics[height=13.5ex, width=19ex]{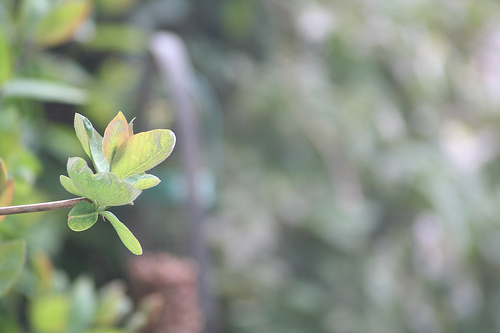}}
\\[-1.5ex]
& \small{Wishful} & \small{I wish I could have a life as easy as a plant.}  \\
& \small{Money-minded} & \small{This plant is probably worth a lot of money}  \\
& \small{Critical} & \small{the leaf is ruining the picture}  \\
& \small{Humble} & \small{This plant is a symbol of life in humble opinion. Just gorgeous!}  \\
& \small{Paranoid} & \small{If you eat this leaf it definetly will not poison you. Or will it...}  \\[1.4ex] 
\hline
%Image 90076977
\\[-1.8ex]
\multirow{ 5}{*}{\includegraphics[height=13.5ex, width=19ex]{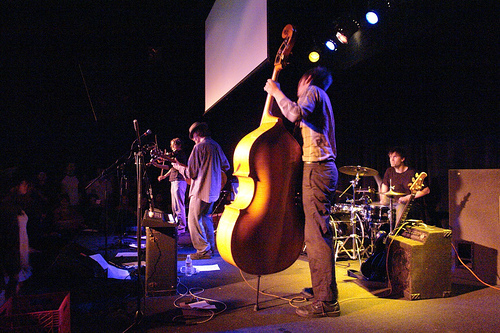}}
\\[-1.5ex]
& \small{Romantic} & \small{This valentine concert is for lovers.}  \\
& \small{Boyish} & \small{It's always fun to get down and jam with the boys!}  \\
& \small{Creative} & \small{musician performing a song of theirs}  \\
& \small{Sweet} & \small{oh what lovely young musicians}  \\
& \small{Money-minded} & \small{I wonder how much the musicians have in student loan debt.}  \\[1.4ex] 
\hline
%Image 67115060
\\[-1.8ex]
\multirow{ 5}{*}{\includegraphics[height=13.5ex, width=19ex]{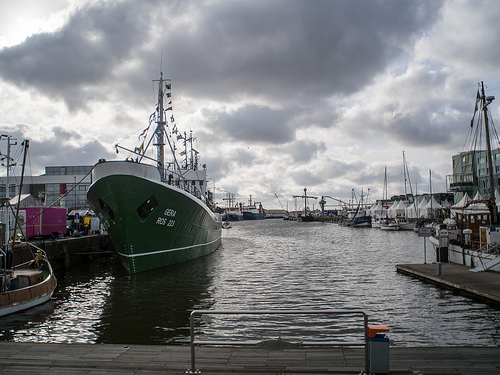}}
\\[-1.5ex]
& \small{Skeptical} & \small{I wonder why the ships are all parked further down the deck.}  \\
& \small{Paranoid} & \small{I hope those ships don't sink}  \\
& \small{Happy} & \small{Look how beautiful the port is at this time of day! :)}  \\
& \small{Arrogant} & \small{Those boats don't need to be docked at this time of night}  \\
& \small{Humble} & \small{We are so lucky to have these boats available locally}  \\[1.4ex] 
\hline
\end{tabular}
\end{adjustbox}
\caption{More example predictions from our best \textsc{\ourmodel} model on the \textsc{\personacaptions} validation set.}
\end{table*}

\end{document}